\documentclass[runningheads]{llncs}

\usepackage{accv}

\usepackage{accvabbrv}

\usepackage{graphicx}
\usepackage{booktabs}
\usepackage{wrapfig}
\usepackage[accsupp]{axessibility}  % Improves PDF readability for those with disabilities.

\usepackage[pagebackref,breaklinks,colorlinks,citecolor=accvblue]{hyperref}
\usepackage{hyperref}

\usepackage{orcidlink}
\usepackage{booktabs}
\usepackage{multirow}
\usepackage[table]{xcolor} 
\newcommand{\best}[1]{\textbf{#1}} 
\newcommand{\second}[1]{\underline{#1}} 
\newcommand{\ours}{\rowcolor{TealBlue!20}}

\usepackage{wrapfig}
\newcommand{\wrapfill}{\WFclear}

\usepackage{algorithm}
\usepackage{algpseudocode}

\begin{document}

\title{TRACE: Time-Adaptive Residual Attention Control with Content–Style Decomposition for Training-Free Diffusion Style Transfer} 

\titlerunning{TRACE: Time-Adaptive Attention Control for Style Transfer}

\author{Khoan Duc Le\inst{1, 2}\and
Ngoc Kim Tran\inst{1, 2}\and
Nhat Minh Le\inst{1, 2}\and
An Thanh Tran\inst{1, 2}\and
Toan Viet Nguyen\inst{1, 2}\and
An Khanh Lay\inst{1, 2}\and
Son Thai Tran\inst{1, 2}\and
Hoang Minh Pham\inst{1, 2}
}
\authorrunning{D.K Le et al.}

\institute{Faculty of Information Technology, University of Science,\\ Ho Chi Minh City, Vietnam \\
% \email{\{ldkhoan, ttson, pmhoang\}@fit.hcmus.edu.vn} \\
% \email{\{23120062, 23120067, 23120017, 23120096, 23120204\}@student.hcmus.edu.vn}
\and
Vietnam National University, Ho Chi Minh City, Vietnam
}

\maketitle

%


\begin{abstract}
Reference-guided style transfer aims to preserve the semantic structure of a content image while transferring the visual appearance of a style reference. Recent diffusion-based methods achieve impressive stylization quality by exploiting strong pretrained generative priors. However, training-free approaches still face a difficult trade-off among style fidelity, content preservation, and content leakage. Direct style injection may unintentionally transfer semantic content from the style image, while fixed guidance schedules often ignore the time- and state-dependent nature of diffusion sampling. To address these limitations, we propose \textbf{TRACE}, a training-free diffusion style transfer framework with \textbf{T}ime-adaptive \textbf{R}esidual \textbf{A}ttention Control and \textbf{C}ontent-Style d\textbf{E}composition. TRACE first performs offline CLIP-based subspace analysis to separate content and style directions from paired data. During inference, it removes content-related components from the style reference and style-related components from the content reference to reduce leakage. It then injects style information through residual cross-attention and applies uncertainty-aware guidance to adapt the guidance signal at each denoising step. Experiments show that TRACE achieves a favorable trade-off between stylization and preservation. Compared with optimal-control-based baselines, TRACE substantially improves style fidelity (+17.28 CSD and +34.10 SRA). While, compared with stylization methods, it better preserves content structure (+12.80 DINO, +5.52 CLIP-I, and -8.19 LPIPS) and reduces directional semantic leakage by 29.5\% in DCL. Our code is publicly available at \href{https://github.com/pixelchemy-research/TRACE}{\textcolor{magenta}{\nolinkurl{TRACE}}}.

\end{abstract}

\section{Introduction}

\label{sec:intro}
Image style transfer aims to synthesize an image that preserves the semantic structure of a content image while adopting the visual appearance of a style reference. Early neural style transfer methods demonstrated that deep features can separate and recombine content and style representations \cite{gatys2016image}, while arbitrary style transfer methods improved efficiency through feature-statistics matching, whitening-and-coloring transforms, attention matching, and transformer-based fusion \cite{huang2017adain, li2017wct, park2019sanet, deng2022stytr2}. More recently, diffusion models have provided strong generative priors for high-fidelity image synthesis and editing \cite{ho2020ddpm, song2021ddim, rombach2022ldm}, motivating diffusion-based style transfer methods based on inversion, attention manipulation, reference feature injection, or training-free guidance \cite{zhang2023inst, kwon2023diffuseit, chung2024styleid, wang2024instantstyle}.

Despite this progress, training-free diffusion style transfer still faces a difficult trade-off among style fidelity, content preservation, and content leakage suppression. A style reference image is not a pure style signal: it often contains colors, textures, object identity, semantic content, and spatial layout. Directly injecting such features may therefore transfer undesired content from the style reference to the generated image, a problem observed in recent reference-guided stylization and personalization methods \cite{wang2024instantstyle, rout2025rbmodulation}. In addition, strong attention replacement or feature injection can improve stylization but may overwrite the original content trajectory and distort object structure \cite{chung2024styleid}. Finally, most guidance strategies still rely on fixed or heuristic schedules 
\cite{ho2022cfg, bansal2023universal, yu2023freedom, ye2024tfg}, despite the time-dependent and state-dependent nature of diffusion sampling. Recent uncertainty-aware diffusion studies show that uncertainty can identify unreliable denoising regions by estimating generation confidence or score variance under perturbations \cite{bayesdiff, aleatoric}. This motivates using uncertainty as a state-dependent signal to adapt guidance along the reverse trajectory.

Motivated by these observations, we propose \textbf{TRACE}, a training-free framework for diffusion-based style transfer. TRACE treats stylization as a controlled denoising process instead of a static feature-transfer operation. First, TRACE performs an offline CLIP-based content-style subspace decomposition from paired content-style data. At inference time, the learned subspaces remove content-related components from the style reference and style-related components from the content reference, reducing reference-induced leakage before sampling. Second, TRACE introduces residual attention control, which injects the cleaned style signal as an additive residual rather than directly replacing attention features. This preserves the content trajectory while allowing style cues to interact with content geometry. Third, TRACE formulates style guidance as a time-dependent control signal and adapts it according to the uncertainty of the current denoising state, strengthening corrective guidance in unstable states while avoiding excessive perturbation in stable ones. Our contributions are summarized as follows:
\enlargethispage{\baselineskip}
\begin{itemize}
    \item We propose a CLIP-based content-style decomposition mechanism that reduces reference-induced content leakage before diffusion sampling.
    \item A residual style injection is introduced to control the injection of style information additively without directly overwriting the content generation trajectory.
    \item We also formulate training-free diffusion style transfer as a time-dependent control problem and develop an uncertainty-aware adaptive guidance strategy for state-dependent stylization control.
    \item Extensive quantitative, qualitative, and ablation studies show that TRACE achieves a favorable balance among style fidelity, content preservation, and leakage suppression.
\end{itemize}

\section{Related Work}

\subsection{Training-free Diffusion Style Transfer}

Diffusion models have become powerful generative priors for image synthesis and editing through iterative denoising \cite{ho2020ddpm, song2021ddim, song2020score}, while Latent Diffusion Models improve efficiency by operating in a compressed latent space with flexible attention-based conditioning \cite{rombach2022ldm}. These advances have motivated diffusion-based style transfer methods that exploit pretrained diffusion models. InST \cite{zhang2023inst} learns a style representation from a single painting through inversion, and DiffuseIT \cite{kwon2023diffuseit} performs image-guided translation with disentangled content and style representations. More recent training-free methods avoid fine-tuning by manipulating attention or reference features. StyleID \cite{chung2024styleid} injects style information through self-attention feature manipulation, while InstantStyle \cite{wang2024instantstyle} emphasizes the need to decouple style and content information in feature space.

Despite their effectiveness, training-free reference-guided methods still face a difficult trade-off between stylization strength and content preservation. A style image often contains not only colors, textures, and artistic patterns, but also object identity, spatial layout, and scene semantics. Directly injecting such entangled features may cause content leakage, where undesired semantic structures from the style reference appear in the output \cite{wang2024instantstyle, rout2025rbmodulation}. Existing methods reduce this issue through feature statistics, attention matching, or feature decoupling \cite{huang2017adain, park2019sanet, kwon2023diffuseit, wang2024instantstyle, attenst}. However, most training-free methods do not explicitly remove content-related components from the style reference before sampling. TRACE addresses this limitation by learning CLIP-based content and style subspaces, which are used to clean both references before attention-based style injection.

\subsection{Attention Control and Adaptive Guidance}

Attention layers provide an effective interface for controlling diffusion generation. Prompt-to-Prompt \cite{hertz2022prompt} shows that cross-attention maps are closely related to spatial layout, and subsequent works exploit attention features for editing, personalization, and style transfer \cite{mokady2023null, tumanyan2023plug, chung2024styleid}. However, direct attention replacement or key-value injection can overwrite the content trajectory, especially when the style reference contains strong semantic structures. This motivates residual attention control, where style information is injected additively while preserving the original content generation path.

Guidance is another important mechanism for steering diffusion sampling. Classifier-free guidance \cite{ho2022cfg} modifies the denoising trajectory to improve condition adherence, while training-free guidance methods use external objectives or energy functions without updating model weights \cite{bansal2023universal, yu2023freedom, chung2023dps, ye2024tfg}. However, most guidance strategies rely on fixed scales or heuristic schedules, which do not fully account for the time-dependent and state-dependent nature of reverse diffusion. Recent works interpret diffusion sampling as trajectory optimization or stochastic optimal control \cite{rout2025rbmodulation, azangulov2025adaptive}, suggesting that guidance should be treated as a dynamic control signal. Motivated by this view, TRACE introduces uncertainty-aware time-dependent guidance that adapts the control strength according to the stability of each denoising state.

\section{Preliminaries}

\subsection{Diffusion Sampling}

Diffusion models generate samples by iteratively reversing a noising process. 
However, many guidance objectives are naturally defined on clean images rather than noisy intermediate states. Given a noisy latent $x_t$ at timestep $t$, a pretrained denoising network $\epsilon_{\theta}(x_t,t)$ can be used to estimate the corresponding clean latent under DDIM sampling \cite{song2021ddim}:
\begin{equation}
    \hat{x}_{0|t}
    =
    \frac{
        x_t - \sqrt{1-\bar{\alpha}_t}\epsilon_{\theta}(x_t,t)
    }{
        \sqrt{\bar{\alpha}_t}
    }.
    \label{eq:clean_estimate}
\end{equation}
Here, $\bar{\alpha}_t$ denotes the cumulative noise schedule. 
The estimated clean latent $\hat{x}_{0|t}$ serves as an intermediate prediction of the final sample and provides a convenient representation for computing style objectives and guiding the denoising trajectory.

\subsection{Optimal Control Problem}
\label{pre:optima_control}
Optimal control (OC) provides a mathematical framework for finding an optimal strategy to steer a dynamical system while minimizing a cost functional. Starting from an initial state $x_0 \in \mathbb{R}^{d}$, consider the following controlled continuous-time dynamics:
\begin{equation}
    dx_t^u
    =
    \left(
    b(x_t^u,t)
    +
    \sigma_t u(x_t^u,t)
    \right)dt
    +
    \sigma_t dB_t,
    \quad
    x_0^u=x_0,
    \label{eq:controlled_sde}
\end{equation}
where $b:\mathbb{R}^{d}\times[0,1]\rightarrow\mathbb{R}^{d}$ is the drift term, $\sigma_t \in \mathbb{R}$ is the diffusion coefficient, $B_t$ denotes Brownian motion, and 
$u:\mathbb{R}^{d}\times[0,1]\rightarrow\mathbb{R}^{d}$ is the additive control term.

The corresponding optimal control problem aims to find $u$ that minimizes
\begin{equation}
\begin{aligned}
    \min_{u\in\mathcal{U}}
    \quad
    &
    \mathbb{E}
    \left[
    \int_0^1
    \left(
    \frac{1}{2}
    \|u(x_t^u,t)\|^2
    +
    f(x_t^u,t)
    \right)dt
    +
    g(x_1^u)
    \right]  \\
    \mathrm{s.t.}
    \quad
    &
    dx_t^u
    =
    \left(
    b(x_t^u,t)
    +
    \sigma_t u(x_t^u,t)
    \right)dt
    +
    \sigma_t dB_t,
    \quad
    x_0^u=x_0 .
\end{aligned}
\label{eq:oc_objective}
\end{equation}
Here, $f$ is the running cost and $g$ is the terminal cost. 
This class of optimal control problems has been extensively studied in both deterministic and stochastic settings. 
Classical analytical tools such as the Hamilton-Jacobi-Bellman (HJB) equation \cite{fleming2012deterministic} and Pontryagin's Maximum Principle (PMP) \cite{pontryagin1987mathematical}, provide necessary and sufficient conditions for optimality.

\subsection{Uncertainty Estimation}

Uncertainty during diffusion sampling can be estimated by measuring the sensitivity of the denoising network around the current state. 
Given the clean latent estimate $\hat{x}_{0|t}$ from \cref{eq:clean_estimate}, $M$ noisy variants are sampled from the forward noising distribution:
\begin{equation}
    \hat{x}_{t}^{(m)}
    \sim
    q(x_t|\hat{x}_{0|t}),
    \quad
    m=1,\ldots,M.
\end{equation}
The denoising network is then evaluated on these perturbed samples:
\begin{equation}
    e_t^{(m)}
    =
    \epsilon_{\theta}(\hat{x}_{t}^{(m)},t).
\end{equation}
Let $E_t$ denote the tensor obtained by stacking the predicted scores
$\{e_t^{(m)}\}_{m=1}^{M}$, and let $\bar{E}_t$ be their mean. 
The timestep uncertainty is estimated as the variance of the predicted scores:
\begin{equation}
    \Sigma_t
    =
    (E_t-\bar{E}_t)^{\top}
    (E_t-\bar{E}_t)
    ,
    \label{eq:uncertainty}
\end{equation}
Large values in $\Sigma_t$ indicate regions or channels where the denoising prediction is unstable under small perturbations. Here, $\Sigma_t$ uses a diagonal approximation, computing uncertainty independently per latent element without cross-element covariance. Thus, $\Sigma_t$ serves as a local reliability measure at timestep $t$. Further implementation details and computational analyses of the uncertainty estimator are provided in Appendix \ref{app:B1} and \ref{app:E4}, respectively.

\enlargethispage{\baselineskip}
\section{Methodology}
\begin{figure}[t]
    \centering
    \includegraphics[width=\linewidth]{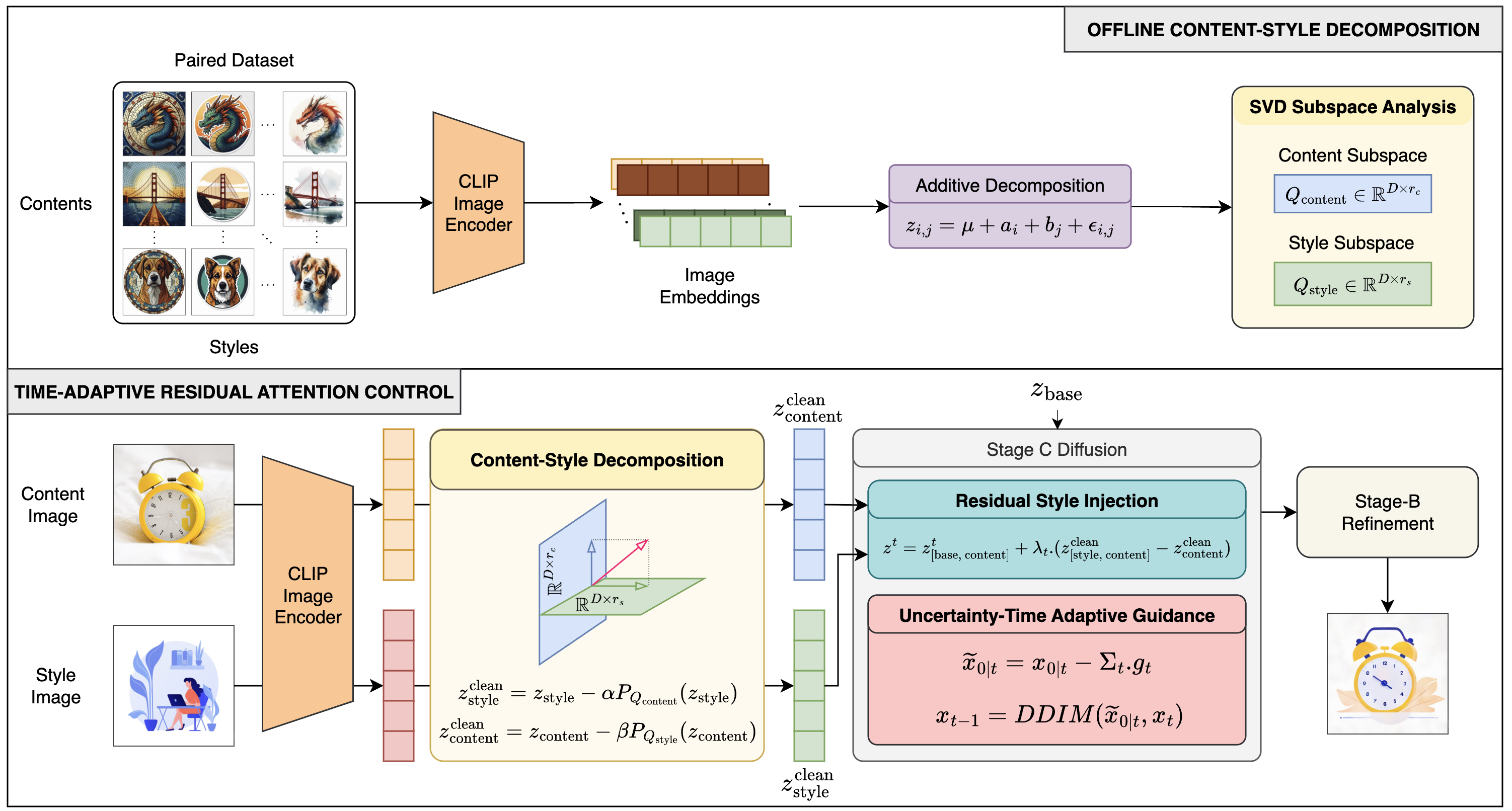}
    \caption{
Overview of our approach (\textbf{TRACE}). TRACE first learns content and style subspaces from paired CLIP embeddings in an offline stage. During inference, it cleans the content and style representations through subspace projection, injects the cleaned style signal via residual style injection, and applies uncertainty-aware time-dependent guidance to control the reverse diffusion trajectory. A detailed TRACE sampling algorithm is provided in Appendix \ref{app:B2}.
}
\label{fig:trace_pipeline}

\end{figure}

\subsection{Overview}

\label{sec:method_overview}

Given a content image $I_c$ and a style reference $I_s$, TRACE generates a stylized image $I_{cs}$ in a training-free manner by reusing a pretrained diffusion model (StableCascade \cite{stablecascade}) without updating its parameters. 
TRACE first performs an offline CLIP-based content-style decomposition, where paired content-style embeddings are used to estimate a content subspace $Q_{\mathrm{content}}$ and a style subspace $Q_{\mathrm{style}}$ via Singular Value Decomposition  (SVD) \cite{Golub1970SingularVD}. 

During inference, these subspaces are used to remove style-related components from the content representation and content-related components from the style representation, yielding cleaned embeddings that reduce reference-induced leakage. The cleaned style signal is then injected through residual attention, preserving the content generation trajectory while transferring style information. Finally, TRACE applies uncertainty-aware adaptive guidance at each denoising step, where the style guidance direction $g_t$ is modulated by the denoising uncertainty $\Sigma_t$ to provide stronger correction in unstable states and more conservative updates in reliable states.
\subsection{Offline Content-Style Decomposition}

A style reference image is usually not a pure style signal. 
Besides color, texture, and brush patterns, it may also contain object identity, layout, and semantic structures. 
Directly using such a reference can therefore introduce undesired content leakage. 
To reduce this issue, TRACE learns content and style subspaces offline from a paired content-style dataset.

Let $\mathcal{D}=\{(I_n,c_n,s_n)\}_{n=1}^{N}$ denote a paired dataset, where $I_n$ is an image, and $c_n$ and $s_n$ are its content and style indices. We extract normalized image embeddings using a frozen CLIP image encoder $E_{\phi}$:
\begin{equation}
    z_n = E_{\phi}(I_n) \in \mathbb{R}^{D}.
\end{equation}
We assume that each embedding can be approximated by an additive content-style model:
\begin{equation}
    z_n = \mu + a_{c_n} + b_{s_n} + \epsilon_n,
    \label{eq:additive_decomposition}
\end{equation}
where $\mu$ is the global mean embedding, $a_{c_n}$ denotes the content-specific effect, $b_{s_n}$ denotes the style-specific effect, and $\epsilon_n$ captures the residual variation.

We estimate the additive effects by solving the following least-squares problem:
\begin{equation}
    \min_{\mu,\{a_i\},\{b_j\}}
    \sum_{n=1}^{N}
    \left\|
    z_n - \mu - a_{c_n} - b_{s_n}
    \right\|_2^2 ,
    \label{eq:additive_lsq}
\end{equation}
where $a_{c_n}$ and $b_{s_n}$ denote the content and style effects of image $n$, respectively. 
Since the content and style effects are coupled, we optimize \cref{eq:additive_lsq} using an alternating iterative procedure. 
At iteration $\ell$, the updates are given by
\begin{equation}
    a_i^{\ell+1}
    =
    \frac{1}{|\Omega_i|}
    \sum_{n\in\Omega_i}
    \left(
    z_n-\mu-b_{s_n}^{\ell}
    \right),
    \quad
    \Omega_i=\{n\mid c_n=i\},
\end{equation}
\begin{equation}
    b_j^{\ell+1}
    =
    \frac{1}{|\Gamma_j|}
    \sum_{n\in\Gamma_j}
    \left(
    z_n-\mu-a_{c_n}^{\ell+1}
    \right),
    \quad
    \Gamma_j=\{n\mid s_n=j\}.
\end{equation}
After each update, we center the effects for identifiability:
\begin{equation}
    a_i^{\ell+1}
    \leftarrow
    a_i^{\ell+1}
    -
    \frac{1}{N_c}
    \sum_{k=1}^{N_c}
    a_k^{\ell+1},
    \quad
    b_j^{\ell+1}
    \leftarrow
    b_j^{\ell+1}
    -
    \frac{1}{N_s}
    \sum_{k=1}^{N_s}
    b_k^{\ell+1}.
\end{equation}
After convergence, we collect the learned content and style effects into matrices
\begin{equation}
    A = [a_1,\ldots,a_{N_c}]^{\top} \in \mathbb{R}^{N_c \times D},
    \quad
    B = [b_1,\ldots,b_{N_s}]^{\top} \in \mathbb{R}^{N_s \times D}.
\end{equation}
We first build the content subspace by applying SVD to the row-normalized content matrix:
\begin{equation}
    \tilde{A} = U_c \Sigma_c V_c^{\top},
    \quad
    Q_{\mathrm{content}} = V_c[:,1:r_c] \in \mathbb{R}^{D \times r_c}.
\end{equation}
To avoid content information leaking into the style basis, we remove the content subspace from the style effects:
\begin{equation}
    B_{\perp}
    =
    \tilde{B}
    -
    \tilde{B}
    Q_{\mathrm{content}}
    Q_{\mathrm{content}}^{\top},
\end{equation}
where $\tilde{B}$ is the row-normalized style-effect matrix. 
We then apply SVD to the content-removed style matrix:
\begin{equation}
    B_{\perp} = U_s \Sigma_s V_s^{\top},
    \quad
    Q_{\mathrm{style}} = V_s[:,1:r_s] \in \mathbb{R}^{D \times r_s}.
\end{equation}
The resulting subspaces $Q_{\mathrm{content}}$ and $Q_{\mathrm{style}}$ are computed once in the offline stage and kept fixed during inference. We provide a pseudocode in Appendix \ref{app:B3}.

At inference time, we extract CLIP embeddings from the content and style images:
\begin{equation}
    z_{\text{content}} = E_{\phi}(I_c),
    \quad
    z_{\text{style}} = E_{\phi}(I_s).
\end{equation}
Using the learned subspaces, we define projection operators:
\begin{equation}
    P_{Q_\text{content}}(z)
    =
    Q_{\text{content}}Q_{\text{content}}^{\top}z,
    \quad
    P_{Q_\text{style}}(z)
    =
    Q_{\text{style}}Q_{\text{style}}^{\top}z.
\end{equation}
To remove undesired entanglement, TRACE cleans the style reference by subtracting its content-related component and cleans the content reference by subtracting its style-related component:
\begin{equation}
    z_{\text{style}}^{\text{clean}}
    =
    z_{\text{style}} - \alpha P_{Q_\text{content}}(z_{\text{style}}),
    \quad
    z_{\text{content}}^{\text{clean}}
    =
    z_{\text{content}} - \beta P_{Q_\text{style}}(z_{\text{content}}),
\end{equation}
where $\alpha$ and $\beta$ control the strength of decomposition. The subspace rank analysis is provided in Appendix \ref{app:C2}. Further analyses of style disentanglement and held-out style-family generalization are provided in Appendix \ref{app:E6}.

\subsection{Residual Style Injection}

After content-style cleaning, TRACE injects the cleaned style signal into the diffusion model by making use of cross-attention. Let $z^{t}_{[\text{base},\text{content}]}$ denote the latent feature at timestep $t$ obtained by cross-attention using the base query and K, V constructed from the self-attention feature and the content representation. This branch mainly preserves the spatial structure of the current diffusion trajectory while injecting content information into the latent representation. 
To extract the stylization effect, TRACE compares two cleaned attention outputs: $z^{\text{clean}}_{[\text{style},\text{content}]}$, obtained by jointly using the cleaned style and content representations, and $z^{\text{clean}}_{\text{content}}$, obtained from the cleaned content representation alone. Their difference gives a residual style signal:
\begin{equation}
    \Delta z^{\text{clean}}
    =
    z^{\text{clean}}_{[\text{style},\text{content}]}
    -
    z^{\text{clean}}_{\text{content}} .
\end{equation}
This residual captures the style information that is injected when the style representation interacts with the content structure. Using both content and style in the attention computation is important because injecting style alone may break the spatial layout or object structure, while the joint content-style attention keeps the stylization aligned with the content geometry.

The final latent feature is updated as
\begin{equation}
    z^{t}
    =
    z^{t}_{[\text{base},\text{content}]}
    +
    \lambda_t
    \Delta z^{\text{clean}},
    \label{eq:residual_injection}
\end{equation}
where $\lambda_t$ controls the strength of residual style injection. To balance the magnitudes of content and style information, we compute $\lambda_t$ according to the relative norm between the base content latent and the residual style signal:
\begin{equation}
    \lambda_t
    =
    \frac{
    \left\| z^{t}_{[\text{base},\text{content}]} \right\|_2
    }{
    \left\| \Delta z^{\text{clean}} \right\|_2 + \epsilon
    } .
\end{equation}
This normalization scales the residual style signal to a comparable magnitude with the content latent, preventing the style residual from being dominated by the content trajectory. As a result, TRACE can inject visible style information while still preserving the spatial structure and semantic content of the input image.

\subsection{Uncertainty-Time Adaptive Guidance}
As discussed above, while existing methods typically rely on a fixed guidance scale, TRACE treats style guidance as an uncertainty-shaped control signal. Following the optimal control formulation in \cref{pre:optima_control}, the aim of this problem is to seek an additional control $u_t^\star$ that minimizes the style distance between the style image and the generated image while keeping the control effort small.

For style transfer, we define the uncertainty-shaped control objective as
\begin{equation}
\begin{aligned}
    \min_{u\in\mathcal{U}}
    \quad
    &
    \mathbb{E}
    \left[
    \int_0^1
    \frac{1}{2}
    u_t^{\top}\Sigma_t^{-1}u_t
    \,dt
    +
    \mathcal{L}_{\mathrm{s}}(x_{0|t}^u; I_s)
    \right],
\end{aligned}
\label{eq:trace_oc_objective}
\end{equation}
where $\mathcal{L}_{\mathrm{s}}$ is the style objective and $\Sigma_t\succ0$ is the uncertainty-shaped control gain at timestep $t$.

Let $\hat{x}_{0|t}$ denote the predicted clean latent at timestep $t$. 
We define the timestep guidance direction as
\begin{equation}
    g_t
    =
    \nabla_{\hat{x}_{0|t}}
    \mathcal{L}_{\mathrm{s}}(\hat{x}_{0|t}; I_s).
    \label{eq:guidance_direction}
\end{equation}
To obtain a tractable update, we locally approximate the style objective around $\hat{x}_{0|t}$. 
For a small control perturbation $u_t$, the first-order Taylor expansion gives
\begin{equation}
    \mathcal{L}_{\mathrm{s}}(\hat{x}_{0|t}+u_t; I_s)
    \approx
    \mathcal{L}_{\mathrm{s}}(\hat{x}_{0|t}; I_s)
    +
    g_t^{\top}u_t .
    \label{eq:local_style_approx}
\end{equation}
Since $\mathcal{L}_{\mathrm{s}}(\hat{x}_{0|t}; I_s)$ is constant with respect to $u_t$, the local one-step control problem becomes
\begin{equation}
    \min_{u_t}
    \quad
    \frac{1}{2}
    u_t^{\top}\Sigma_t^{-1}u_t
    +
    g_t^{\top}u_t ,
    \label{eq:local_control_problem}
\end{equation}
where the first term penalizes the uncertainty-modulated control effort and the second term encourages reducing the style objective. Taking the derivative with respect to $u_t$ yields
\begin{equation}
    \nabla_{u_t}
    \left(
    \frac{1}{2}
    u_t^{\top}\Sigma_t^{-1}u_t
    +
    g_t^{\top}u_t
    \right)
    =
    \Sigma_t^{-1}u_t + g_t .
    \label{eq:local_derivative}
\end{equation}
Setting \cref{eq:local_derivative} to zero yields the closed-form local control:
\begin{equation}
    u_t^\star
    =
    -\Sigma_t g_t .
    \label{eq:closed_form_control}
\end{equation}
This control term is applied to the predicted clean latent $\hat{x}_{0|t}$, producing the uncertainty-guided clean estimate:
\begin{equation}
    \widetilde{x}_{0|t}
    =
    \hat{x}_{0|t}
    +
    u_t^\star
    =
    \hat{x}_{0|t}
    -
    \Sigma_t g_t .
    \label{eq:trace_clean_update}
\end{equation}
The updated clean estimate $\widetilde{x}_{0|t}$ is then used in the DDIM transition to obtain the next latent state:
\begin{equation}
    x_{t-1}
    =
    \mathrm{DDIM}
    \left(
        \widetilde{x}_{0|t},
        x_t
    \right).
    \label{eq:trace_ddim_update}
\end{equation}

This update allows TRACE to strengthen corrective guidance in uncertain states while keeping the update conservative when the denoising trajectory is already stable. The derivation of this objective and the corresponding closed-form control update are provided in Appendix \ref{app:oc_derivation}.

\section{Experimental Setups} 
\subsection{Baselines}
TRACE is compared with two groups of representative training-free diffusion baselines. Stylization methods, including StyleID\cite{chung2024styleid}, AttenST\cite{attenst}, DiffuseIT\cite{kwon2023diffuseit}, and InstantStyle\cite{wang2024instantstyle}, mainly focus on transferring visual style through feature, attention, or reference-guided manipulation. 
OC-based methods, including RB-Modulation\cite{rout2025rbmodulation} and ITOC\cite{chang2025training}, emphasize controlled reverse diffusion and source preservation. Detailed descriptions of all methods and implementation settings of TRACE are provided in Appendix \ref{app:D1} and \ref{app:D2}, respectively.

\subsection{Dataset}
For inference-time style transfer evaluation, we use 15 style reference images from the StyleAligned benchmark \cite{hertz2024style} and 13 content images from DreamBooth \cite{ruiz2023dreambooth}. For the offline content-style decomposition stage, we use the paired content-style data from RB-Modulation \cite{rout2025rbmodulation}. Detailed statistics and analysis of the decomposition dataset are provided in Appendix \ref{app:C1}.

\subsection{Evaluation Metrics}
We evaluate TRACE from three aspects: \textbf{style fidelity}, \textbf{content preservation}, and \textbf{content leakage}. 
\begin{itemize}
    \item Style fidelity is measured by CSD style similarity (CSD) \cite{csd} and Style Retrieval Accuracy (SRA). 
    \item Content preservation is measured by CLIP image similarity (CLIP-I) \cite{clip}, DINO image similarity (DINO) \cite{dino}, and Learned Perceptual Image Patch Similarity (LPIPS) \cite{lpips}.
    \item Content leakage is measured by Directional Content Leakage (DCL). 
\end{itemize}
Detailed metric definitions are provided in Appendix \ref{app:D3}.
\enlargethispage{\baselineskip}
\section{Results}
The parameters for main experiment are selected based on \cref{sec:ablation_study}, while the subspace rank analysis is provided in Appendix \ref{app:C2}.
\subsection{Qualitative Results}

\begin{figure}[tb]
    \centering
    \includegraphics[width=\linewidth]{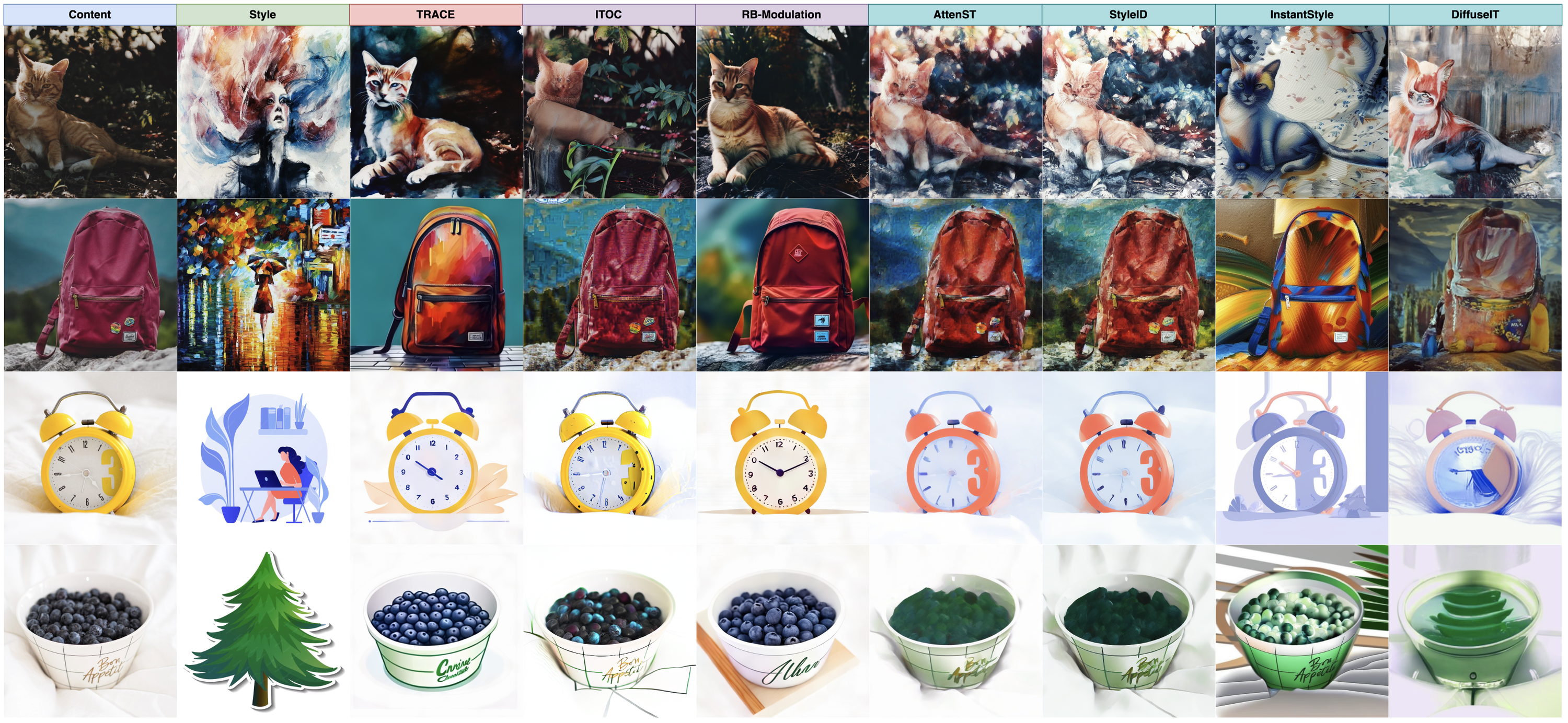}
    \caption{Qualitative comparison with training-free style transfer methods. Red, purple, and cyan boxes denote TRACE, OC-based methods, and stylization methods, respectively. More visualizations are provided in Appendix~\ref{app:E4}.} 
    \label{fig:qualitative_results}
\end{figure}
\Cref{fig:qualitative_results} shows qualitative comparisons between TRACE and existing training-free baselines. OC-based methods such as ITOC and RB-Modulation generally preserve the content structure and produce visually stable images, but their stylization strength is relatively weak, especially when the target style requires more than simple color transfer. In contrast, stylization-oriented methods such as AttenST, StyleID, InstantStyle, and DiffuseIT often produce stronger style effects, but they may distort the object structure or introduce semantic content from the style reference. This is mainly because strong attention manipulation or AdaIN can act as a direct information transfer between the content and style images, which improves style fidelity but also increases the risk of content leakage. TRACE achieves a more balanced result. It preserves the main object layout and spatial structure while injecting visible style patterns into the generated image. The advantage is particularly clear for styles that involve structural changes (samples 3, 4) rather than only color shifts (samples 1, 2). These results demonstrate that the proposed content-style decomposition, residual style injection, and uncertainty-time adaptive guidance jointly improve stylization quality without severely damaging the content structure. Additional qualitative comparisons and diverse stylization results are provided in Appendix \ref{app:E8}.

\subsection{Quantitative Results}
\begin{table*}[t]
  \scriptsize
  \caption{
  Quantitative comparison with stylization and OC-based methods. $\uparrow$ indicates higher is better, while $\downarrow$ indicates lower is better. Best results are shown in bold, and second-best results are underlined within each group.
  }
  \label{tab:main_results}
  \centering
  \resizebox{\textwidth}{!}{
  \begin{tabular}{@{}ccccccc@{}}
    \toprule
    \multirow{2}{*}{\textbf{Model}}
    & \multicolumn{2}{c}{\textbf{Style Fidelity}}
    & \multicolumn{3}{c}{\textbf{Content Preservation}}
    & \multicolumn{1}{c}{\textbf{Leakage}} \\
    \cmidrule(lr){2-3} \cmidrule(lr){4-6} \cmidrule(l){7-7}
    & CSD $\uparrow$ & SRA $\uparrow$
    & DINO $\uparrow$ & CLIP-I $\uparrow$ & LPIPS $\downarrow$
    & DCL $\downarrow$ \\
    \midrule

    \multicolumn{7}{@{}l}{\textbf{\textit{Stylization Methods}}} \\
    \midrule

    \ours TRACE
    & 34.920
    & 69.744
    & \best{65.543}
    & \best{77.271}
    & \second{39.532}
    & \best{21.568} \\

    StyleID
    & \best{44.998}
    & \best{93.333}
    & 60.798
    & 74.418
    & 42.633
    & 26.415 \\

    AttenST
    & \second{42.876}
    & \second{88.718}
    & \second{62.115}
    & \second{74.757}
    & \best{38.538}
    & \second{25.699} \\

    DiffuseIT
    & 26.468
    & 62.051
    & 42.908
    & 70.109
    & 51.083
    & 38.193 \\

    InstantStyle
    & 39.485
    & 87.692
    & 45.148
    & 67.740
    & 58.646
    & 32.048 \\

    \midrule
    \multicolumn{7}{@{}l}{\textbf{\textit{OC-based Methods}}} \\
    \midrule

    \ours TRACE
    & \best{34.920}
    & \best{69.744}
    & 65.543
    & 77.271
    & 39.532
    & 21.568 \\

    RB-Modulation
    & \second{18.369}
    & 34.359
    & \best{75.212}
    & \best{85.168}
    & \second{34.406}
    & \best{14.526} \\

    ITOC
    & 16.918
    & \second{36.923}
    & \second{70.826}
    & \second{83.996}
    & \best{34.396}
    & \second{19.866} \\

    \bottomrule
  \end{tabular}
  }
\end{table*}

\begin{wrapfigure}{r}{0.48\textwidth}
    \centering
    % \vspace{-0.1cm}
    \includegraphics[width=0.48\textwidth]{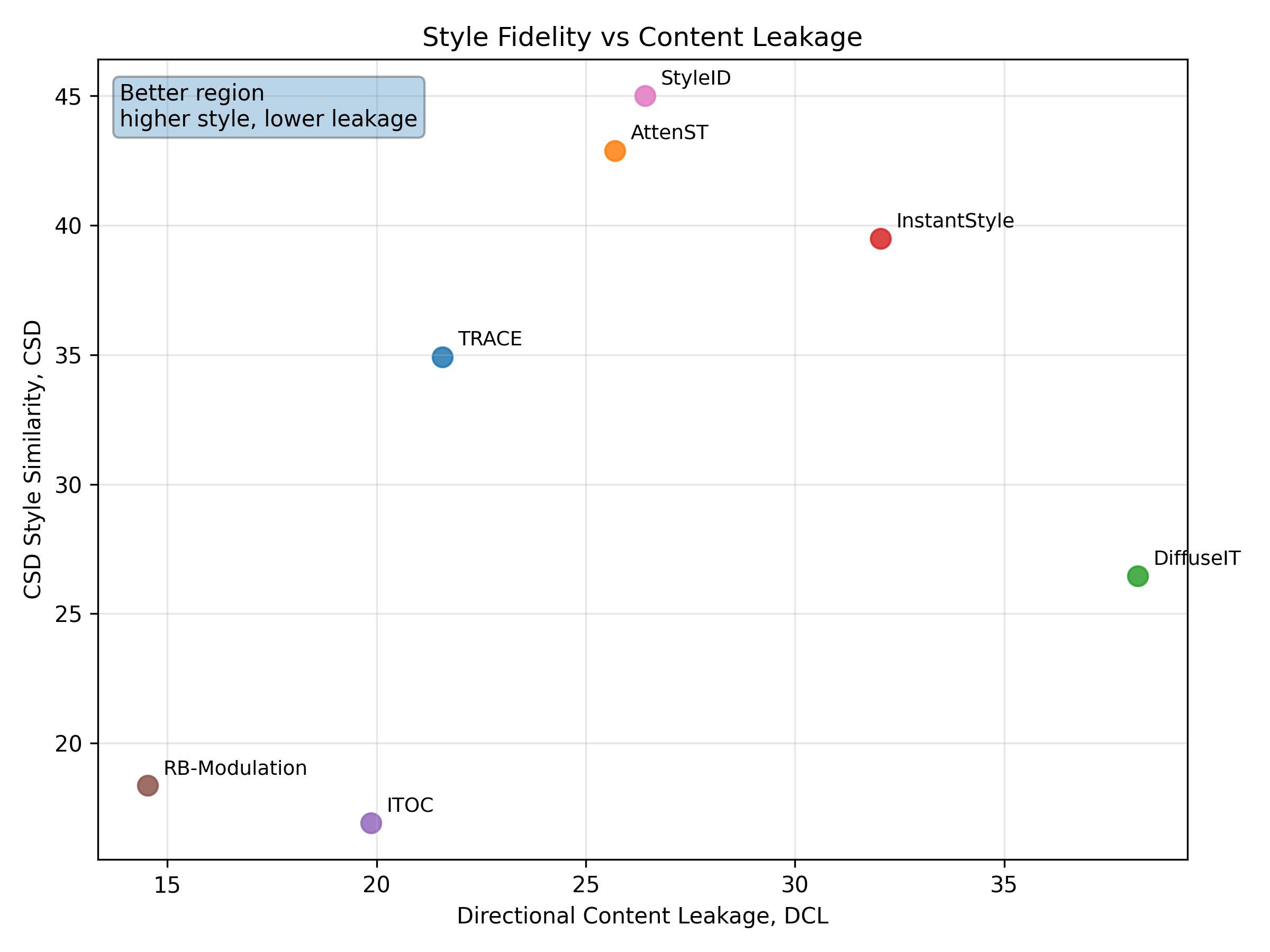}
    \caption{ Style fidelity and content leakage trade-off. The $y$-axis shows CSD, and the $x$-axis shows DCL. TRACE achieves effective stylization while maintaining lower leakage than most stylization baselines. Additional Pareto trade-off analyses are provided in Appendix \ref{app:E7}.}
    \label{fig:style_leakage_tradeoff}
    \vspace{-0.3cm}
\end{wrapfigure}

\Cref{fig:style_leakage_tradeoff} visualizes the trade-off between style fidelity and content leakage. Stylization methods such as StyleID, AttenST, and InstantStyle achieve higher SRA scores, but they also introduce larger DCL values, indicating stronger semantic leakage from the style reference. In contrast, OC-based methods such as RB-Modulation and ITOC obtain low DCL scores, but their CSD values remain lower, suggesting insufficient stylization. TRACE achieves a more balanced trade-off: it sacrifices a small amount of style fidelity compared with the strongest stylization baselines, but better preserves content structure and substantially reduces directional content leakage.
\wrapfill

\paragraph{\textbf{Comparison with Stylization Methods:}}\Cref{tab:main_results} compares TRACE with representative diffusion-based stylization methods. StyleID and AttenST achieve the highest style scores in terms of CSD and SRA, showing that strong attention-based stylization can effectively transfer visual appearance. However, these methods provide weaker content preservation than TRACE, as reflected by lower DINO and CLIP-I scores. DiffuseIT and InstantStyle also show noticeable degradation in content consistency, especially on DINO, CLIP-I, and LPIPS. In contrast, TRACE achieves the best content preservation on DINO and CLIP-I, while maintaining competitive style fidelity. Although TRACE does not obtain the highest style scores, it significantly reduces semantic leakage from the style reference, achieving the lowest DCL among all compared stylization methods. This indicates that TRACE better preserves the semantic structure of the content image while avoiding undesired transfer of object identity or layout from the style image.
\enlargethispage{\baselineskip}

\paragraph{\textbf{Comparison with OC-based Methods:}}\Cref{tab:main_results} compares our approach with optimal-control-based methods, including RB-Modulation and ITOC. These methods achieve strong content preservation, as shown by higher DINO and CLIP-I scores and lower LPIPS and DCL values. This indicates that OC-based methods are effective at maintaining the source structure and reducing semantic leakage from the style reference. However, their stylization strength is limited, with substantially lower CSD and SRA scores than TRACE. In contrast, TRACE significantly improves style fidelity over both RB-Modulation and ITOC, achieving the best CSD and SRA scores among this group. Although TRACE has slightly weaker content-preservation and leakage scores than these conservative OC-based baselines, it provides much stronger stylization while still maintaining reasonable content consistency. These results show that TRACE provides a better style-content trade-off for reference-guided style transfer.

\section{Ablation Study}
\label{sec:ablation_study}

\begin{table*}[t]
  \caption{
  Unified ablation study of TRACE.
  E0 denotes the full TRACE model, while E1-E8 denote different ablation settings. $D_s$ denotes the content-style decomposition strength, where $0$, $0.6$, and $1$ indicate no decomposition, soft decomposition, and full decomposition, respectively. $R$ denotes residual attention control, where $\checkmark$ indicates residual style injection and $\times$ indicates the original cross-attention injection without residual control. $G$ denotes the guidance strategy, where Ada., No, C1, and C2 indicate uncertainty-time adaptive guidance, no adaptive guidance, constant guidance with $\lambda=1$, and constant guidance with $\lambda=2$, respectively. $\rho$ denotes the noise level, where $0.5$, $0.7$, and $1.0$ correspond to low-noise, ours, and high-noise settings. $\uparrow$ indicates higher is better, while $\downarrow$ indicates lower is better.
  }
  \label{tab:ablation_all}
  \centering
  \scriptsize
  \resizebox{\textwidth}{!}{
  \begin{tabular}{@{}lcccccccccc@{}}
    \toprule
    \multirow{2}{*}{Exp}
    & \multirow{2}{*}{$D_s$}
    & \multirow{2}{*}{$R$}
    & \multirow{2}{*}{$G$}
    & \multirow{2}{*}{$\rho$}
    & \multicolumn{2}{c}{\textbf{Style Fidelity}}
    & \multicolumn{3}{c}{\textbf{Content Preservation}}
    & \multicolumn{1}{c}{\textbf{Leakage}} \\
    \cmidrule(lr){6-7} \cmidrule(lr){8-10} \cmidrule(l){11-11}
    &
    &
    &
    &
    & CSD $\uparrow$ & SRA $\uparrow$
    & DINO $\uparrow$ & CLIP-I $\uparrow$ & LPIPS $\downarrow$
    & DCL $\downarrow$ \\
    \midrule

    \ours E0
    & 0.6
    & $\checkmark$
    & Ada.
    & 0.7
    & 34.920
    & 69.744
    & 65.543
    & 77.271
    & 39.532
    & 21.568 \\

    \midrule

    E1
    & 0
    & $\checkmark$
    & Ada.
    & 0.7
    & 53.020
    & 90.769
    & 51.586
    & 69.538
    & 44.768
    & 34.098 \\

    E2
    & 1
    & $\checkmark$
    & Ada.
    & 0.7
    & 26.848
    & 56.410
    & 69.295
    & 78.048
    & 38.002
    & 17.634 \\

    \midrule

    E3
    & 0.6
    & $\times$
    & Ada.
    & 0.7
    & 17.874
    & 32.821
    & 75.491
    & 84.742
    & 34.489
    & 14.441 \\

    \midrule

    E4
    & 0.6
    & $\checkmark$
    & No
    & 0.7
    & 20.111
    & 34.359
    & 65.539
    & 79.208
    & 37.444
    & 19.585 \\

    E5
    & 0.6
    & $\checkmark$
    & C1
    & 0.7
    & 33.383
    & 68.718
    & 64.465
    & 77.431
    & 38.823
    & 21.157 \\

    E6
    & 0.6
    & $\checkmark$
    & C2
    & 0.7
    & 42.195
    & 78.974
    & 60.507
    & 74.344
    & 43.206
    & 24.049 \\

    \midrule

    E7
    & 0.6
    & $\checkmark$
    & Ada.
    & 1.0
    & 55.837
    & 94.872
    & 38.774
    & 64.528
    & 68.934
    & 43.805 \\

    E8
    & 0.6
    & $\checkmark$
    & Ada.
    & 0.5
    & 27.574
    & 56.410
    & 74.785
    & 82.139
    & 28.345
    & 14.642 \\

    \bottomrule
  \end{tabular}
  }
\end{table*}
\subsection{Effect of Content-Style Decomposition}

We first analyze the effect of content-style decomposition by comparing E0 with E1 and E2 in \cref{tab:ablation_all}. Without decomposition, E1 achieves the highest style scores, but its content preservation drops significantly and its DCL becomes much higher. This indicates that directly using the style reference can transfer strong visual appearance, but it also introduces semantic leakage from the style image. In contrast, full decomposition in E2 improves content preservation and reduces leakage, but it also removes useful style information, leading to weaker stylization. The soft decomposition used in TRACE provides a better balance between these two extremes, maintaining effective style transfer while reducing content leakage.

\subsection{Effect of Residual Style Injection}
We then evaluate the effect of residual style injection by comparing E0 with E3. The original cross-attention variant in E3 preserves content well, as shown by higher DINO and CLIP-I scores and lower leakage. However, its style fidelity is much weaker, suggesting that the style signal is insufficiently injected into the denoising trajectory. By contrast, the residual style injection in TRACE substantially improves CSD and SRA while still keeping reasonable content preservation. This confirms that adding a content-aligned stylization residual is more effective than conservative attention fusion for reference-guided style transfer. The detail of original cross-attention is provided in Appendix \ref{app:E1}.

\subsection{Effect of Time-adaptive Guidance}
To study the effect of guidance strength, we compare E0 with E4-E6. From no adaptive guidance in E4 to constant guidance with $\lambda=1$ in E5, stronger guidance improves stylization, indicating that an appropriate control signal is necessary for transferring style effectively. However, further increasing the fixed guidance scale to $\lambda=2$ in E6 leads to overly aggressive stylization, which weakens content preservation and increases leakage. This trend shows that a fixed guidance value is difficult to balance across different denoising states. In contrast, TRACE uses uncertainty-time adaptive guidance to adjust the control strength along the reverse trajectory, achieving a better balance between style fidelity and content preservation. Additional analyses of timestep-wise uncertainty and adaptive guidance are provided in Appendices \ref{app:E2} and \ref{app:E5}.

\subsection{Effect of Noise-level Scheduling}

To study the effect of the starting noise level, we compare different values of $\rho$. Instead of starting from pure noise, TRACE starts sampling from a fixed intermediate noising step, which helps preserve the pose and spatial structure of the content image. When $\rho=1.0$, the trajectory starts from a highly noisy state, it allows stronger stylization but causes noticeable content distortion and leakage. When $\rho=0.5$, more content information is retained, leading to better pose preservation but weaker style transfer. We therefore use $\rho=0.7$ in TRACE, which provides a better balance between preserving the content pose and enabling sufficient stylization. Additional visual results under different noise levels are provided in Appendix \ref{app:E3}.

\section{Conclusion}

In this paper, we presented TRACE, a training-free diffusion framework for reference-guided style transfer. TRACE integrates content-style decomposition, residual style injection, and uncertainty-time adaptive guidance to balance style fidelity, content preservation, and leakage suppression. Experiments show that TRACE better preserves content structure and reduces semantic leakage compared with stylization methods while achieving stronger stylization than OC-based baselines. However, TRACE still has limitations when the style reference contains strong structural patterns or shape-dominant elements. In such cases, the transferred style may affect the pose, contour, or fine geometry of the content object, leading to imperfect shape preservation. In future work, we will explore more explicit structure- and pose-preserving mechanisms to further protect the spatial layout of the content image during style injection.

\subsection*{Acknowledgement}

This research is funded by University of Science, VNU-HCM, Vietnam under grant number CNTT 2025-12.

% \clearpage
\bibliographystyle{splncs04}
\bibliography{main}

\clearpage
\appendix
\renewcommand{\theHsection}{appendix.\Alph{section}}
\renewcommand{\theHsubsection}{appendix.\Alph{section}.\arabic{subsection}}

\addcontentsline{toc}{section}{Appendix}
\section*{Appendix}
\vspace{0.3cm}

\noindent
\small
\begin{tabular}{@{}p{0.95\textwidth}r@{}}

  \ref{app:oc_derivation} \quad Derivation of Uncertainty-aware Optimal Control \dotfill
  & \pageref{app:oc_derivation} \\

  \qquad \ref{app:A1} \quad From Stochastic Optimal Control to Uncertainty-shaped Control \dotfill
  & \pageref{app:A1} \\

  \qquad \ref{app:A2} \quad Closed-form Solution of the Local Control Problem \dotfill
  & \pageref{app:A2} \\

  \qquad \ref{app:A3} \quad TRACE Guidance Update \dotfill
  & \pageref{app:A3} \\[0.25cm]

  \ref{app:algorithm} \quad Pseudocode \dotfill
  & \pageref{app:algorithm} \\

  \qquad \ref{app:B1} \quad Uncertainty Estimation \dotfill
  & \pageref{app:B1} \\

  \qquad \ref{app:B2} \quad TRACE Sampling \dotfill
  & \pageref{app:B2} \\

  \qquad \ref{app:B3} \quad Offline Content-Style Decomposition \dotfill
  & \pageref{app:B3} \\[0.25cm]

  \ref{app:decomposition} \quad Additional Details of Content-Style Decomposition \dotfill
  & \pageref{app:decomposition} \\

  \qquad \ref{app:C1} \quad Construction of Paired Content-Style Dataset \dotfill
  & \pageref{app:C1} \\

  \qquad \ref{app:C2} \quad Rank Selection \dotfill
  & \pageref{app:C2} \\[0.25cm]

  \ref{app:experimental_setup} \quad Detailed Experimental Setups \dotfill
  & \pageref{app:experimental_setup} \\

  \qquad \ref{app:D1} \quad Baseline Details \dotfill
  & \pageref{app:D1} \\

  \qquad \ref{app:D2} \quad Configuration of TRACE \dotfill
  & \pageref{app:D2} \\

  \qquad \ref{app:D3} \quad Evaluation Metrics \dotfill
  & \pageref{app:D3} \\[0.25cm]

  \ref{app:result_analysis} \quad Extended Result Analysis \dotfill
  & \pageref{app:result_analysis} \\

  \qquad \ref{app:E1} \quad Residual Style Injection vs. Original Attention \dotfill
  & \pageref{app:E1} \\

  \qquad \ref{app:E2} \quad Timestep-wise Uncertainty Analysis \dotfill
  & \pageref{app:E2} \\

  \qquad \ref{app:E3} \quad Analysis of the Starting Noise Step \dotfill
  & \pageref{app:E3} \\

  \qquad \ref{app:E4} \quad
  Uncertainty Sample Sensitivity and Computational Cost \dotfill
  & \pageref{app:E4} \\

  \qquad \ref{app:E5} \quad
  Adaptive Guidance vs. Handcrafted Schedules \dotfill
  & \pageref{app:E5} \\

  \qquad \ref{app:E6} \quad
  Style Disentanglement and Held-out Style Generalization \dotfill
  & \pageref{app:E6} \\

  \qquad \ref{app:E7} \quad
  Trade-off Analysis \dotfill
  & \pageref{app:E7} \\

  \qquad \ref{app:E8} \quad Additional Qualitative Results \dotfill
  & \pageref{app:E8} \\

\end{tabular}
\normalsize

\vspace{0.5cm}
\noindent Our appendix is organized into five sections:
\begin{itemize}

    \item \Cref{app:oc_derivation} derives the uncertainty-aware optimal control formulation, its closed-form local controller, and the resulting TRACE guidance update.

    \item \Cref{app:algorithm} provides pseudocode for uncertainty estimation, TRACE sampling, and the offline content-style decomposition procedure.

    \item \Cref{app:decomposition} provides additional details of the CLIP-based content-style decomposition, including paired dataset construction and subspace rank selection.

    \item \Cref{app:experimental_setup} describes the experimental setup, including baseline methods, TRACE configurations, and evaluation metrics.

    \item \Cref{app:result_analysis} presents extended experimental analyses, including residual style injection, timestep-wise uncertainty, starting noise level, uncertainty efficiency, adaptive guidance, style disentanglement and generalization, Pareto trade-offs, and additional qualitative results.

\end{itemize}

\newpage

\section{Derivation of Uncertainty-aware Optimal Control}
\label{app:oc_derivation}
This section derives the uncertainty-aware guidance update used in TRACE. We first formulate diffusion guidance as an uncertainty-shaped stochastic optimal control problem, derive the corresponding global optimality condition, and then obtain a tractable local controller used in our sampling procedure.

\subsection{From Stochastic Optimal Control to Uncertainty-shaped Control}
\label{app:A1}

Following the optimal control formulation in \cref{pre:optima_control}, we view guided reverse diffusion as a controlled stochastic process:
\begin{equation}
    dx_t^u
    =
    \left(
        b(x_t^u,t)
        +
        \sigma_t u_t
    \right)dt
    +
    \sigma_t dB_t,
    \qquad
    x_T^u=x_T ,
    \label{eq:app_controlled_sde}
\end{equation}
where $b(x_t^u,t)$ is the base reverse drift, $\sigma_t$ is the diffusion coefficient, and $u_t$ is an additional control term. The goal is to steer the reverse trajectory toward the target style while keeping the control effort small.

A standard stochastic optimal control objective can be written as
\begin{equation}
    \min_{u\in\mathcal{U}}
    \mathbb{E}
    \left[
        \int_T^1
        \frac{1}{2}
        \|u_t\|_2^2
        \,dt
        +
        \mathcal{L}_{\mathrm{s}}
        \left(
            x_1^u;
            I_s
        \right)
    \right],
    \label{eq:app_standard_oc}
\end{equation}
subject to \cref{eq:app_controlled_sde}. 
However, the isotropic penalty $\frac{1}{2}\|u_t\|_2^2$ treats all latent directions equally, although diffusion states have different reliability across timesteps and latent dimensions.

To account for this heterogeneity, TRACE replaces the isotropic control penalty with an uncertainty-shaped Mahalanobis cost:
\begin{equation}
    \frac{1}{2}
    \|u_t\|_2^2
    \quad
    \longrightarrow
    \quad
    \frac{1}{2}
    u_t^\top
    \Sigma_t^{-1}
    u_t ,
    \label{eq:app_uncertainty_penalty}
\end{equation}
where $\Sigma_t\succ0$ is an uncertainty-shaped control gain. Large values in $\Sigma_t$ correspond to uncertain directions, making the corresponding entries of $\Sigma_t^{-1}$ smaller. Thus, control along uncertain directions is penalized less, while control along stable directions remains more conservative.

The uncertainty-shaped optimal control problem becomes
\begin{equation}
    \min_{u\in\mathcal{U}}
    \mathbb{E}
    \left[
        \int_T^1
        \frac{1}{2}
        u_t^\top
        \Sigma_t^{-1}
        u_t
        \,dt
        +
        \mathcal{L}_{\mathrm{s}}
        \left(
            x_1^u;
            I_s
        \right)
    \right],
    \label{eq:app_uncertainty_oc}
\end{equation}
subject to \cref{eq:app_controlled_sde}. In the diagonal approximation, the uncertainty-shaped gain is defined element-wise as
\begin{equation}
    \Sigma_{t,i}
    =
    1+\bar{U}_{t,i},
    \qquad i=1,\ldots,d,
    \label{eq:app_sigma_construction}
\end{equation}
where the unit offset preserves the baseline guidance, while $\bar{U}_t$ provides uncertainty-dependent amplification along individual latent directions.

To derive the global optimality condition, we define the Hamiltonian:
\begin{equation}
    H(x_t,u_t,p_t,t)
    =
    p_t^\top
    \left(
        b(x_t,t)
        +
        \sigma_t u_t
    \right)
    +
    \frac{1}{2}
    u_t^\top
    \Sigma_t^{-1}
    u_t ,
    \label{eq:app_hamiltonian}
\end{equation}
where $p_t$ denotes the adjoint state. According to Pontryagin's Maximum Principle, the optimal control satisfies
\begin{equation}
    u_t^\star
    =
    \arg\min_{u_t}
    H(x_t^\star,u_t,p_t^\star,t).
    \label{eq:app_pmp_min}
\end{equation}
Taking the derivative of the Hamiltonian with respect to $u_t$ gives
\begin{equation}
    \nabla_{u_t}H
    =
    \sigma_t p_t^\star
    +
    \Sigma_t^{-1}u_t .
    \label{eq:app_hamiltonian_derivative}
\end{equation}
Setting the derivative to zero yields
\begin{equation}
    \Sigma_t^{-1}u_t^\star
    =
    -
    \sigma_t p_t^\star .
    \label{eq:app_global_stationary}
\end{equation}
Therefore, the global optimal control takes the form
\begin{equation}
    u_t^\star
    =
    -
    \Sigma_t
    \sigma_t
    p_t^\star .
    \label{eq:app_global_control}
\end{equation}

This global solution shows that uncertainty directly shapes the geometry of the optimal control. Larger uncertainty increases the corresponding entries of $\Sigma_t$, allowing stronger corrective updates in unreliable directions. However, computing $p_t^\star$ requires solving a coupled state-adjoint system over the whole reverse trajectory, which is intractable for high-dimensional diffusion sampling. Therefore, TRACE adopts a local one-step approximation.

\subsection{Closed-form Solution of the Local Control Problem}
\label{app:A2}

Instead of solving the full global control problem, TRACE derives a local controller at each denoising step. Let $\hat{x}_{0|t}$ denote the predicted clean latent at timestep $t$. Since style objectives are naturally evaluated on clean latent estimates, TRACE applies the control directly to $\hat{x}_{0|t}$.

The timestep guidance direction is defined as
\begin{equation}
    g_t
    =
    \nabla_{\hat{x}_{0|t}}
    \mathcal{L}_{\mathrm{s}}
    \left(
        \hat{x}_{0|t};
        I_s
    \right).
    \label{eq:app_guidance_direction}
\end{equation}
For a small local perturbation $u_t$, we approximate the style objective by first-order Taylor expansion:
\begin{equation}
    \mathcal{L}_{\mathrm{s}}
    \left(
        \hat{x}_{0|t}+u_t;
        I_s
    \right)
    \approx
    \mathcal{L}_{\mathrm{s}}
    \left(
        \hat{x}_{0|t};
        I_s
    \right)
    +
    g_t^\top u_t .
    \label{eq:app_style_taylor}
\end{equation}
Since the first term is constant with respect to $u_t$, the local one-step control problem becomes
\begin{equation}
    \min_{u_t}
    \quad
    \frac{1}{2}
    u_t^\top
    \Sigma_t^{-1}
    u_t
    +
    g_t^\top u_t .
    \label{eq:app_local_control_problem}
\end{equation}

Because $\Sigma_t\succ0$, the objective in \cref{eq:app_local_control_problem} is strictly convex and has a unique minimizer. Taking the derivative with respect to $u_t$ gives
\begin{equation}
    \nabla_{u_t}
    \left(
        \frac{1}{2}
        u_t^\top
        \Sigma_t^{-1}
        u_t
        +
        g_t^\top u_t
    \right)
    =
    \Sigma_t^{-1}u_t
    +
    g_t .
    \label{eq:app_local_derivative}
\end{equation}
Setting the derivative to zero yields
\begin{equation}
    \Sigma_t^{-1}u_t^\star
    +
    g_t
    =
    0 .
    \label{eq:app_local_stationary}
\end{equation}
Thus,
\begin{equation}
    \Sigma_t^{-1}u_t^\star
    =
    -
    g_t .
    \label{eq:app_local_solve}
\end{equation}
Multiplying both sides by $\Sigma_t$, we obtain the closed-form local control:
\begin{equation}
    u_t^\star
    =
    -
    \Sigma_t g_t .
    \label{eq:app_closed_form_control}
\end{equation}

This local controller can be viewed as a tractable one-step approximation of the global uncertainty-shaped control in \cref{eq:app_global_control}. The adjoint direction in the global solution is replaced by the instantaneous style gradient $g_t$, while the uncertainty-shaped gain $\Sigma_t$ is retained to modulate the control strength.

\subsection{TRACE Guidance Update}
\label{app:A3}

TRACE applies the closed-form local control in \cref{eq:app_closed_form_control} to the predicted clean latent:
\begin{equation}
    \widetilde{x}_{0|t}
    =
    \hat{x}_{0|t}
    +
    u_t^\star
    =
    \hat{x}_{0|t}
    -
    \Sigma_t g_t .
    \label{eq:app_trace_clean_update}
\end{equation}
The updated clean estimate $\widetilde{x}_{0|t}$ is then used in the DDIM transition:
\begin{equation}
    x_{t-1}
    =
    \mathrm{DDIM}
    \left(
        \widetilde{x}_{0|t},
        x_t
    \right).
    \label{eq:app_trace_ddim_update}
\end{equation}

Therefore, the TRACE update is a local uncertainty-shaped control step applied during reverse diffusion. 
When the denoising state is uncertain, $\Sigma_t$ increases the effective guidance strength and allows stronger corrective updates. When the state is reliable, the update remains more conservative, helping preserve the content trajectory.

\newpage
\section{Pseudocode}
\label{app:algorithm}
\subsection{Uncertainty Estimation}
\label{app:B1}
The uncertainty approximation procedure is summarized in \cref{alg:uncertainty_estimation}. The $M$ re-noised samples are evaluated in a single batch, and their element-wise variance provides a diagonal uncertainty estimate. After percentile normalization and $\tanh$ bounding, this adds only one extra denoising-network forward pass per timestep.

\begin{algorithm}[H]
\caption{Uncertainty Estimation}
\label{alg:uncertainty_estimation}

\begin{algorithmic}[1]
\Require Current latent $x_t$, timestep $t$, denoising network
$\epsilon_{\theta}$, cumulative noise schedule $\bar{\alpha}_t$,
number of perturbations $M$, small constant $\varepsilon$
\Ensure Normalized uncertainty map $\bar{U}_t$

\State Estimate the clean latent:
\[
    \hat{x}_{0|t}
    =
    \frac{
        x_t - \sqrt{1-\bar{\alpha}_t}\epsilon_{\theta}(x_t,t)
    }{
        \sqrt{\bar{\alpha}_t}
    } .
\]

\For{$m=1,\ldots,M$}
    \State Sample Gaussian noise
    $\xi^{(m)} \sim \mathcal{N}(0,\mathbf{I})$.

    \State Re-noise the clean estimate:
    \[
        \hat{x}_{t}^{(m)}
        =
        \sqrt{\bar{\alpha}_t}\hat{x}_{0|t}
        +
        \sqrt{1-\bar{\alpha}_t}\xi^{(m)} .
    \]

    \State Compute the perturbed denoising prediction:
    \[
        e_t^{(m)}
        =
        \epsilon_{\theta}(\hat{x}_{t}^{(m)},t).
    \]
\EndFor

\State Compute the element-wise uncertainty across perturbations:
\[
    U_t^{\mathrm{raw}}
    =
    \operatorname{Var}_{m}
    \left(
        \left\{
        e_t^{(m)}
        \right\}_{m=1}^{M}
    \right),
\]

\State Compute the $90$-th percentile:
\[
    q_{0.9,t}
    =
    \operatorname{Quantile}_{0.9}
    \left(
        U_t^{\mathrm{raw}}
    \right).
\]

\State Normalize and bound the uncertainty:
\[
    \bar{U}_t
    =
    \tanh
    \left(
        \frac{
            U_t^{\mathrm{raw}}
        }{
            q_{0.9,t}+\varepsilon
        }
    \right).
\]

\State \Return $\bar{U}_t$
\end{algorithmic}
\end{algorithm}

\subsection{TRACE Sampling}
\label{app:B2}
The complete TRACE sampling procedure, including content-style cleaning, residual style injection, uncertainty estimation, and adaptive guidance, is summarized in \cref{alg:trace_sampling}.

\begin{algorithm}[H]
\caption{TRACE Sampling}
\label{alg:trace_sampling}

\begin{algorithmic}[1]
\Require Content image $I_c$, style image $I_s$, pretrained diffusion model $\epsilon_{\theta}$, image encoder $\mathcal{E}$, image decoder $\mathcal{D}$, CLIP image encoder $E_{\phi}$, learned subspaces $Q_{\mathrm{content}}$ and $Q_{\mathrm{style}}$, decomposition strengths $\alpha,\beta$, starting noise level $\rho$, denoising timesteps $\{T_{\rho},\ldots,1\}$, number of local guidance steps $N$, uncertainty parameters $(M,\varepsilon)$
\Ensure Stylized image $I_{cs}$

\State Extract CLIP embeddings:
\[
    z_{\mathrm{content}} = E_{\phi}(I_c),
    \qquad
    z_{\mathrm{style}} = E_{\phi}(I_s).
\]

\State Clean the content and style representations:
\[
    z_{\mathrm{content}}^{\mathrm{clean}}
    =
    z_{\mathrm{content}}
    -
    \beta
    P_{Q_{\mathrm{style}}}(z_{\mathrm{content}}),
    \qquad
    z_{\mathrm{style}}^{\mathrm{clean}}
    =
    z_{\mathrm{style}}
    -
    \alpha
    P_{Q_{\mathrm{content}}}(z_{\mathrm{style}}).
\]

\State Encode the content image into the latent space and initialize the starting latent:
\[
    x_0^{c}
    =
    \mathcal{E}(I_c), \qquad 
     x_{T_{\rho}}
    =
    \sqrt{\bar{\alpha}_{T_{\rho}}}x_0^{c}
    +
    \sqrt{1-\bar{\alpha}_{T_{\rho}}}\xi,
    \qquad
    \xi\sim\mathcal{N}(0,I).
\]
This initialization starts from a noised content latent instead of pure random noise, helping preserve the pose and spatial structure of the content image.

\For{$t=T_{\rho},\ldots,1$}

    \State Predict the noise and estimate the predicted clean latent:
    \[
        \epsilon_t
        =
        \epsilon_{\theta}(x_t,t), \qquad
        \hat{x}_{0|t}
        =
        \frac{
            x_t
            -
            \sqrt{1-\bar{\alpha}_t}\epsilon_t
        }{
            \sqrt{\bar{\alpha}_t}
        }.
    \]

    \State Apply residual style injection:
    \[
        \Delta z_t^{\mathrm{clean}}
        =
        z_{[\mathrm{style},\mathrm{content}]}^{\mathrm{clean}}
        -
        z_{\mathrm{content}}^{\mathrm{clean}}, \qquad 
        z^t
        =
        z^t_{[\mathrm{base},\mathrm{content}]}
        +
        \lambda_t
        \Delta z_t^{\mathrm{clean}}.
    \]

    \State Estimate the normalized element-wise uncertainty map $\bar{U}_t$
using \cref{alg:uncertainty_estimation}.

\State Construct the diagonal uncertainty-shaped control gain $\Sigma_t$
using \cref{eq:app_sigma_construction}.

    \State Initialize the locally guided clean latent:
    \[
        \widetilde{x}_{0|t}^{(0)}
        =
        \hat{x}_{0|t}.
    \]

    \For{$n=1,\ldots,N$}
        \State Compute the style guidance direction:
        \[
            g_t^{(n)}
            =
            \nabla_{\widetilde{x}_{0|t}^{(n-1)}}
            \mathcal{L}_{\mathrm{s}}
            \left(
                \widetilde{x}_{0|t}^{(n-1)};
                I_s
            \right).
        \]

        \State Apply uncertainty-aware control:
        \[
            \widetilde{x}_{0|t}^{(n)}
            =
            \widetilde{x}_{0|t}^{(n-1)}
            -
            \Sigma_t g_t^{(n)} .
        \]
    \EndFor

    \State Set the final guided clean estimate and update the next latent using DDIM:
    \[
        \widetilde{x}_{0|t}
        =
        \widetilde{x}_{0|t}^{(N)}, \qquad  
        x_{t-1}
        =
        \mathrm{DDIM}
        \left(
            \widetilde{x}_{0|t},
            x_t
        \right).
    \]
\EndFor

\State Decode the final latent:
\[
    I_{cs}
    =
    \mathcal{D}(x_0).
\]

\State \Return $I_{cs}$

\end{algorithmic}
\end{algorithm}

\subsection{Offline Content-Style Decomposition}
The offline decomposition procedure is summarized in \cref{alg:offline_decomposition}.
\label{app:B3}
\begin{algorithm}[H]
\caption{Offline Content-Style Decomposition}
\label{alg:offline_decomposition}

\begin{algorithmic}[1]
\Require Paired dataset $\mathcal{D}=\{(I_n,c_n,s_n)\}_{n=1}^{N}$, CLIP encoder $E_{\phi}$, ranks $r_c,r_s$, iterations $L$
\Ensure Content subspace $Q_{\mathrm{content}}$ and style subspace $Q_{\mathrm{style}}$

\State Extract and normalize CLIP embeddings:
\[
    z_n \leftarrow \frac{E_{\phi}(I_n)}{\|E_{\phi}(I_n)\|_2},
    \quad n=1,\ldots,N .
\]
\State Compute global mean $\mu=\frac{1}{N}\sum_{n=1}^{N}z_n$ and initialize $a_i^0=0$, $b_j^0=0$.

\For{$\ell=0,\ldots,L-1$}
    \State Update content effects:
    \[
        a_i^{\ell+1}
        \leftarrow
        \frac{1}{|\Omega_i|}
        \sum_{n\in\Omega_i}
        (z_n-\mu-b_{s_n}^{\ell}),
        \quad
        \Omega_i=\{n:c_n=i\}.
    \]
    \State Center $a_i^{\ell+1} \leftarrow a_i^{\ell+1}-\frac{1}{N_c}\sum_{k=1}^{N_c}a_k^{\ell+1}$.

    \State Update style effects:
    \[
        b_j^{\ell+1}
        \leftarrow
        \frac{1}{|\Gamma_j|}
        \sum_{n\in\Gamma_j}
        (z_n-\mu-a_{c_n}^{\ell+1}),
        \quad
        \Gamma_j=\{n:s_n=j\}.
    \]
    \State Center $b_j^{\ell+1} \leftarrow b_j^{\ell+1}-\frac{1}{N_s}\sum_{k=1}^{N_s}b_k^{\ell+1}$.
\EndFor

\State Form effect matrices:
\[
    A=[a_1,\ldots,a_{N_c}]^\top,
    \quad
    B=[b_1,\ldots,b_{N_s}]^\top .
\]

\State Row-normalize $A$ and $B$ to obtain $\widetilde{A}$ and $\widetilde{B}$.
\State Compute content subspace:
\[
    \widetilde{A}=U_c\Sigma_cV_c^\top,
    \quad
    Q_{\mathrm{content}}=V_c[:,1:r_c].
\]

\State Remove content directions from style effects:
\[
    B_{\perp}
    =
    \widetilde{B}
    -
    \widetilde{B}Q_{\mathrm{content}}Q_{\mathrm{content}}^\top .
\]

\State Compute style subspace:
\[
    B_{\perp}=U_s\Sigma_sV_s^\top,
    \quad
    Q_{\mathrm{style}}=V_s[:,1:r_s].
\]

\State \Return $Q_{\mathrm{content}}, Q_{\mathrm{style}}$
\end{algorithmic}
\end{algorithm}

\newpage
\section{Additional Details of Content-Style Decomposition}
\label{app:decomposition}

\subsection{Construction of Paired Content-Style Dataset}
\label{app:C1}
\begin{figure}[H]
\vspace{-20pt}
    \centering
    \includegraphics[width=0.8\linewidth]{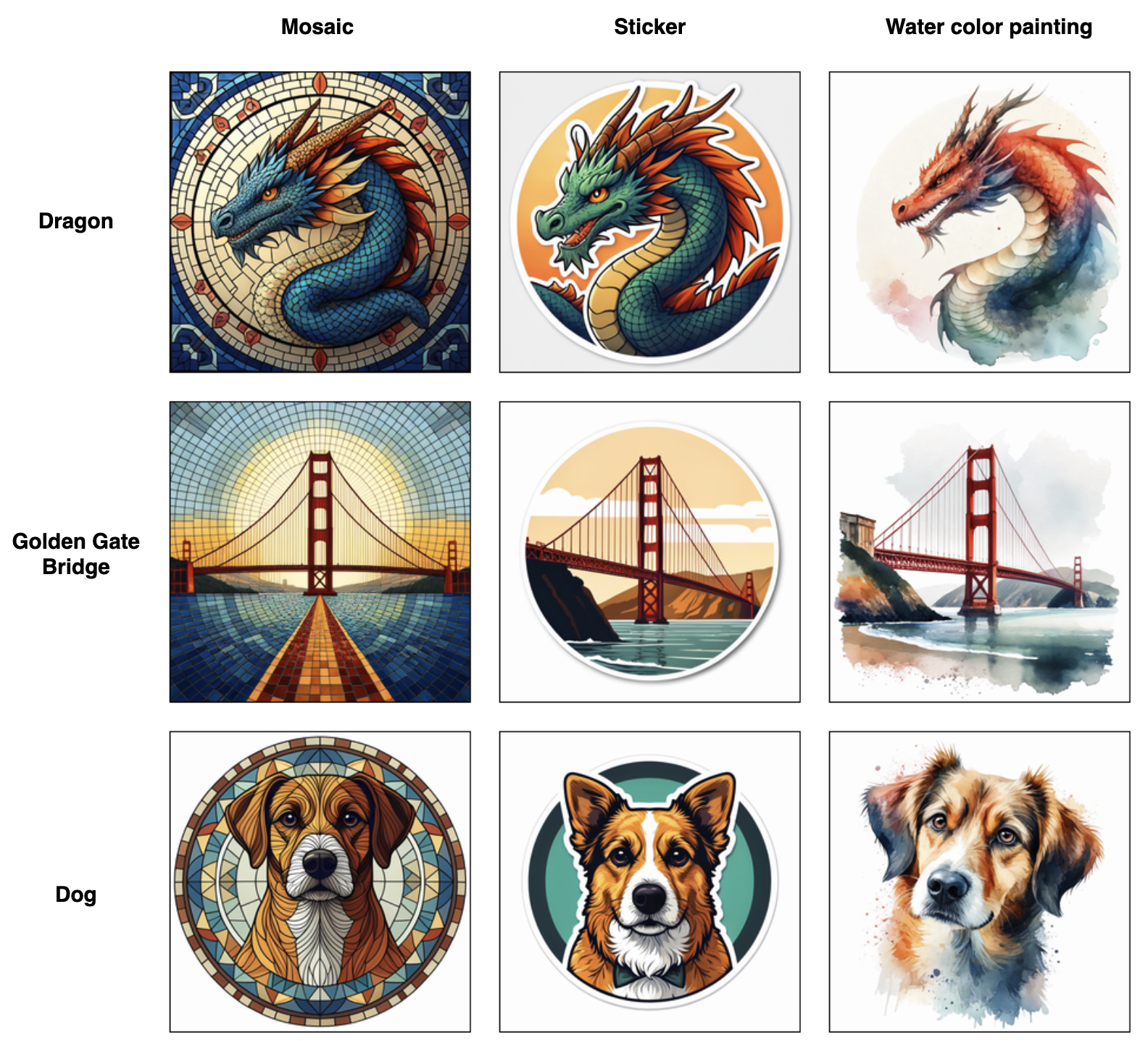}
    \caption{Representative paired content-style samples used for decomposition analysis. Rows show content identities and columns show style references.}
    \label{fig:sample_data_cs}
\vspace{-20pt}
\end{figure}
To estimate the offline content and style subspaces, we construct a paired content-style dataset from 350 content images and 100 style images. Instead of exhaustively combining all content and style references, we randomly sample 10 style images for each content image, producing 3,500 stylized pairs in total. Each generated pair is associated with a content index and a style index, allowing us to analyze content-specific and style-specific variations in the CLIP embedding space. We split the 3,500 paired samples into 3,150 pairs for offline subspace estimation and 350 held-out pairs for decomposition analysis and validation. No parameters of the pretrained diffusion model or the CLIP image encoder are updated during this process. Moreover, the paired data used for subspace construction and decomposition analysis is kept separate from the inference-time evaluation pairs. Therefore, TRACE remains training-free with respect to the diffusion backbone and uses the offline dataset only to estimate reusable content and style directions.

The content set covers common objects, animals, people, vehicles, buildings, landscapes, food, and fantasy objects. The style set covers illustration, painting, 3D rendering, logo, poster, photography, abstract art, and printmaking styles. Representative examples are shown in \cref{tab:content_style_samples} and \cref{fig:sample_data_cs}.

\begin{table}[h]
\centering
\caption{
Representative content and style prompts used for constructing the paired content-style dataset.
}
\label{tab:content_style_samples}
\resizebox{\linewidth}{!}{
\begin{tabular}{@{}ll@{}}
\toprule
\textbf{Category} & \textbf{Representative examples} \\
\midrule
Objects 
& A chair; A teapot; A laptop; A vintage camera; A pocket watch \\

Animals 
& A dog; A lion; A fox; A penguin; A panda \\

People and actions 
& A child running in the park; A person dancing; A woman reading in the park; A man skiing down the hill \\

Vehicles 
& A train; A racing car; A bicycle; A helicopter; A sailboat \\

Buildings and scenes 
& A house; A temple; Golden Gate Bridge; A village; A snowy cabin \\

Nature and food 
& Flowers; A sunflower; A watermelon; A bowl with apples; A mushroom \\

Fantasy and symbolic objects 
& A dragon; An elf; A crown; A magician's hat; A crystal ball \\

\midrule
Illustration styles 
& Sticker style; Line drawing style; Cartoon line drawing style; Flat cartoon illustration style \\

Painting styles 
& Watercolor painting style; Oil painting style; Japanese ukiyo-e style; Hawaiian sunset painting style \\

3D and rendering styles 
& 3D rendering style; Glowing 3D rendering style; Realistic 3D render; Enchanted 3D rendering style \\

Poster and logo styles 
& Vintage poster style; Vintage travel poster style; Minimalist round BW logo style; Cafe logo style \\

Abstract and modern styles 
& Abstract geometric style; Digital glitch art style; Psychedelic art style; Neo-futurism style \\

Print and craft styles 
& Woodblock print style; Papercut art style; Origami style; Woodcut print style \\

Photography styles 
& Macro photography style; Professional food photography style; Bokeh photography style;\\

\bottomrule
\end{tabular}
}
\vspace{-15pt}
\end{table}

\subsection{Rank Selection}
\label{app:C2}

\begin{table}[H]
\vspace{-20pt}
  \caption{
  Top-10 retrieval analysis before and after content-style decomposition. Raw denotes retrieval before decomposition, while Clean denotes retrieval after decomposition. SG indicates style gain and CR indicates content reduction. All values are reported in percentage.
  }
  \label{tab:rank_retrieval_analysis}
  \centering
  \scriptsize
  \resizebox{\linewidth}{!}{
  \begin{tabular}{@{}ccccccc@{}}
    \toprule
    $r_c$ & $r_s$
    & Raw Style & Clean Style & SG $\uparrow$
    & Clean Content & CR $\uparrow$ \\
    \midrule
    32  & 8/16/32 & 8.86 & 15.12 & 6.26  & 32.37 & 21.98 \\
    64  & 8/16/32 & 8.86 & 18.55 & 9.69  & 21.84 & 47.37 \\
    96  & 8/16/32 & 8.86 & 20.53 & 11.68 & 14.13 & 65.95 \\
    \ours 128 & 8/16/32 & 8.86 & \textbf{21.13} & \textbf{12.27} & \textbf{9.25} & \textbf{77.70} \\
    \bottomrule
  \end{tabular}
  }
  \vspace{-15pt}
\end{table}

\Cref{tab:rank_retrieval_analysis} shows that content-style decomposition consistently improves style retrieval and reduces content retrieval as the content rank $r_c$ increases. Starting from the same Raw Style value of $8.86\%$, Clean Style increases from $15.12\%$ at $r_c=32$ to $21.13\%$ at $r_c=128$, with the style gain improving from $6.26\%$ to $12.27\%$. At the same time, Clean Content decreases from $32.37\%$ to $9.25\%$, and content reduction increases from $21.98\%$ to $77.70\%$. These results indicate that a larger content subspace more effectively removes content-related information and makes the cleaned representation more style-discriminative. The results are stable across the tested style ranks $r_s \in \{8,16,32\}$, suggesting that the leading style components already capture the dominant style directions. Therefore, we use $r_c=128$ for strong content removal and choose the compact style rank $r_s=8$ for the main experiments.

\section{Detailed Experimental Setups}
\label{app:experimental_setup}

\subsection{Baseline Details}
\label{app:D1}

We compare TRACE with two groups of diffusion-based baselines: Stylization methods and OC-based methods.

\paragraph{\textbf{Stylization methods:}}
This group focuses on transferring visual appearance from a style reference. StyleID\cite{chung2024styleid} transfers style by manipulating diffusion features and attention representations between the content and style images. AttenST\cite{attenst} injects style information through attention-based manipulation. DiffuseIT\cite{kwon2023diffuseit} performs diffusion-based image translation and stylization using iterative guidance. InstantStyle\cite{wang2024instantstyle} uses reference-guided style injection to transfer style information while reducing content interference. 

\paragraph{\textbf{OC-based methods:}}
This group focuses on controlling the reverse diffusion trajectory. RB-Modulation\cite{rout2025rbmodulation} modulates the reverse sampling process to preserve the source image while steering generation toward the target condition. ITOC\cite{chang2025training} formulates diffusion sampling as a trajectory optimization problem and controls the reverse process during inference. 

\subsection{Configuration of TRACE}
\label{app:D2}

The configuration of TRACE used in the main experiments is summarized in \cref{tab:trace_config}.

\begin{table}[H]
\vspace{-20pt}
\centering
\caption{Configuration of TRACE used in the main experiments.}
\label{tab:trace_config}
\footnotesize
\begin{tabular}{@{}lc@{}}
\toprule
\textbf{Component} & \textbf{Setting} \\
\midrule
Diffusion backbone & Stable Cascade \cite{stablecascade} \\
Content rank & $r_c = 128$ \\
Style rank & $r_s = 8$ \\
Decomposition strength & $\alpha = \beta = 0.6$ \\
Residual style injection & $\checkmark$ \\
Adaptive guidance & $\checkmark$ \\
Uncertainty samples & $M = 4$ \\
Local guidance steps & $N = 2$ \\
Starting noise level & $\rho = 0.7$ \\
\bottomrule
\end{tabular}
\vspace{-15pt}
\end{table}

\subsection{Evaluation Metrics}
\label{app:D3}

\subsubsection{Evaluation Metrics}
\label{app:evaluation_metrics}

We evaluate style transfer from three aspects: style fidelity, content preservation, and content leakage. The metrics used in our experiments are summarized in \cref{tab:evaluation_metrics}.  We further provide detailed definitions of SRA and DCL, which are used to evaluate style retrieval accuracy and directional content leakage, respectively.

\begin{table}[H]
\vspace{-20pt}
\centering
\caption{Evaluation metrics used in our experiments.}
\label{tab:evaluation_metrics}
\footnotesize
\begin{tabular}{@{}lll@{}}
\toprule
\textbf{Aspect} & \textbf{Metric} & \textbf{Interpretation} \\
\midrule
Style fidelity 
& CSD $\uparrow$ 
& Style similarity to the target style reference \\

Style fidelity 
& SRA $\uparrow$ 
& Top-1 retrieval accuracy of the target style \\

Content preservation 
& DINO $\uparrow$ 
& Semantic similarity to the content image \\

Content preservation 
& CLIP-I $\uparrow$ 
& Image-level similarity to the content image \\

Content preservation 
& LPIPS $\downarrow$ 
& Perceptual distance from the content image \\

Content leakage 
& DCL $\downarrow$ 
& Semantic leakage from the style reference \\
\bottomrule
\end{tabular}
\vspace{-15pt}
\end{table}

\paragraph{\textbf{Style Retrieval Accuracy (SRA).}}
SRA evaluates whether the generated image can be correctly retrieved as the target style from a style reference gallery. For each generated image $I_{cs}^{i}$, we extract its CSD style embedding and compare it with the CSD embeddings of all style reference images using cosine similarity. The retrieved style is defined as the nearest style reference in the gallery. 
SRA is computed as:
\begin{equation}
    \mathrm{SRA}
    =
    \frac{1}{N}
    \sum_{i=1}^{N}
    \mathbf{1}
    \left[
        \arg\max_{j}
        \cos
        \left(
            \phi_{\mathrm{CSD}}(I_{cs}^{i}),
            \phi_{\mathrm{CSD}}(I_s^{j})
        \right)
        =
        y_s^{i}
    \right],
\end{equation}
where $\phi_{\mathrm{CSD}}$ denotes the CSD style encoder, $I_s^{j}$ denotes a style reference image in the gallery, and $y_s^{i}$ is the target style label of the $i$-th generated image. A higher SRA indicates that the generated image is more recognizable as the target style.

\paragraph{\textbf{Directional Content Leakage (DCL).}}
DCL evaluates whether the generated image moves toward the semantic structure of the style reference. 
We compute DCL in a content/structure feature space using DINO features extracted from grayscale images, which reduces the influence of color and low-level style statistics. Let $f_c$, $f_s$, and $f_g$ denote the normalized DINO features of the content image, style image, and generated image, respectively. 
DCL is defined as:
\begin{equation}
    \mathrm{DCL}
    =
    \operatorname{clip}_{[0,1]}
    \left(
    \frac{
        (f_g - f_c)^\top (f_s - f_c)
    }{
        \|f_s - f_c\|_2^2 + \epsilon
    }
    \right).
\end{equation}
A larger DCL indicates that the generated image moves more strongly from the content image toward the style reference in the structure feature space, suggesting stronger content leakage. Therefore, lower DCL is better.

\section{Extended Result Analysis}
\label{app:result_analysis}

\subsection{Residual Style Injection vs. Original Attention}
\label{app:E1}

\begin{figure}[H]
\vspace{-20pt}
    \centering

    \begin{subfigure}[t]{0.48\linewidth}
        \centering
        \includegraphics[width=\linewidth]{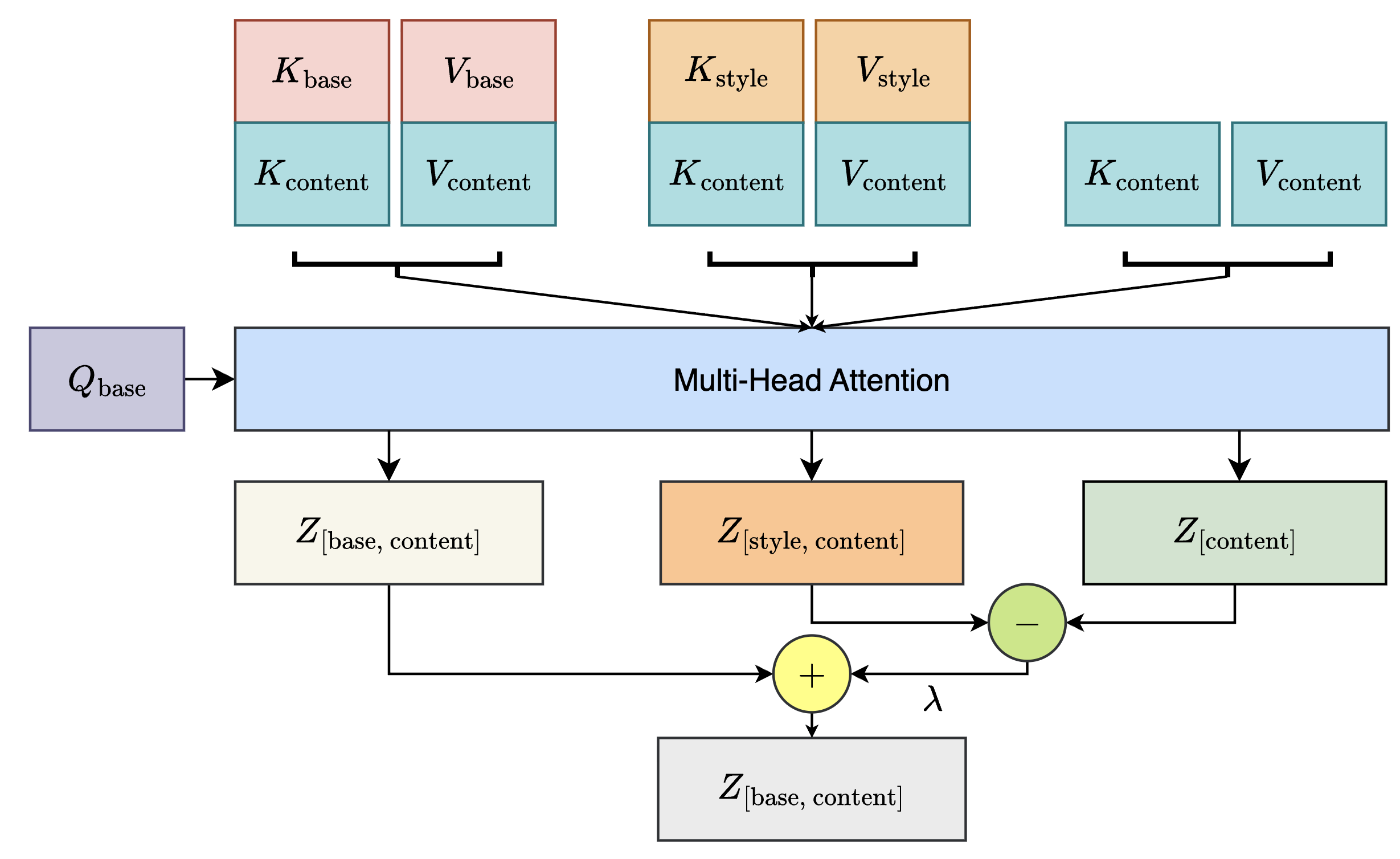}
        \caption{Residual style injection}
        \label{fig:residual_injection}
    \end{subfigure}
    \hfill
    \begin{subfigure}[t]{0.48\linewidth}
        \centering
        \includegraphics[width=\linewidth]{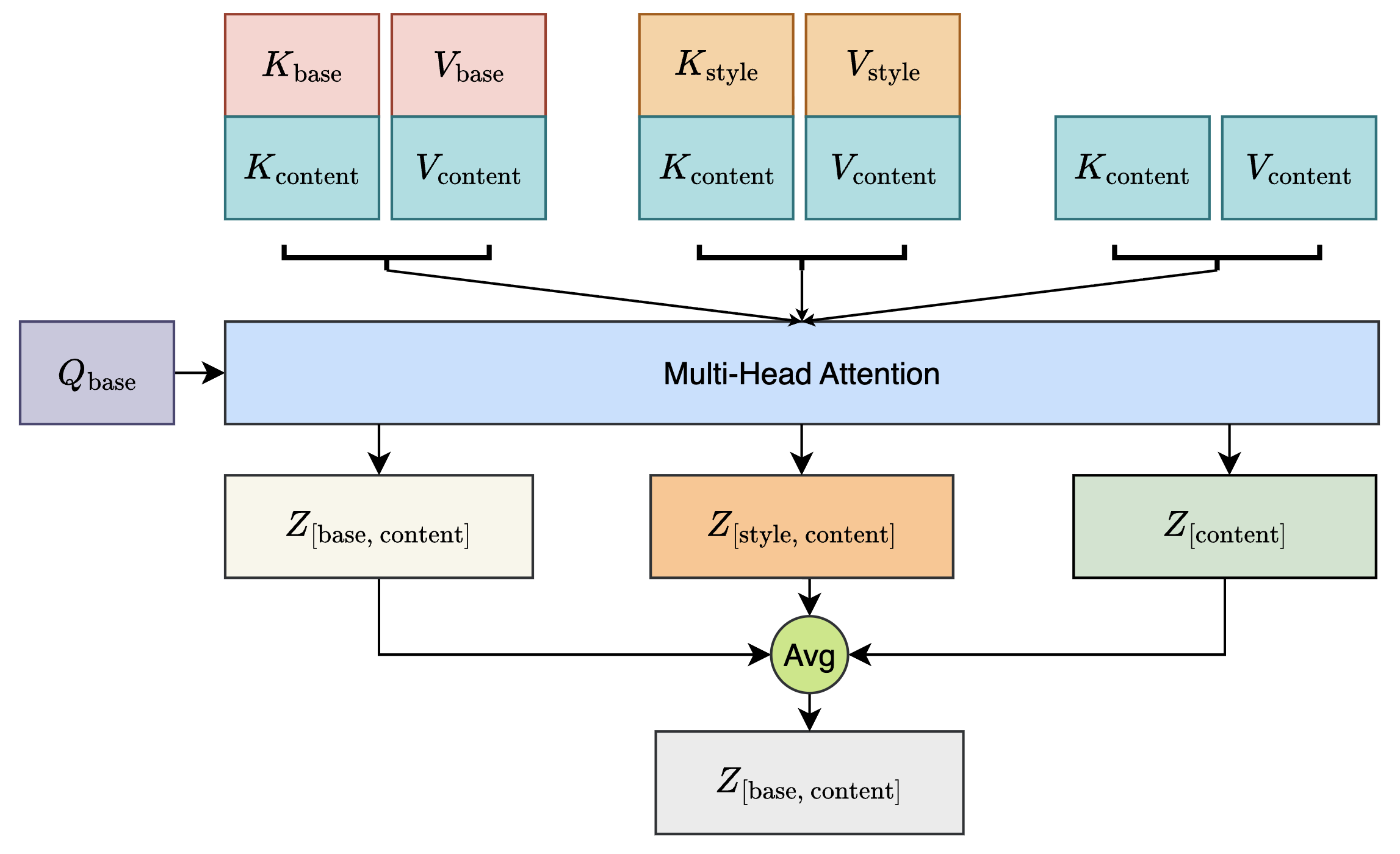}
        \caption{Original attention}
        \label{fig:original_attention}
    \end{subfigure}

    \caption{
    Comparison between residual style injection and original attention.
    }
    \label{fig:residual_vs_original_attention}
    \vspace{-20pt}
\end{figure}

As shown in \cref{fig:residual_vs_original_attention}, the original attention strategy combines multiple attention outputs by simple averaging. Specifically, the outputs obtained from base-content attention, style-content attention, and content-only attention are averaged to produce the final representation. Although this strategy mixes style and content information, it treats all branches equally and may dilute the style signal or disturb the content trajectory.

In contrast, the residual style injection used in TRACE keeps the base-content attention output as the main trajectory and injects style only through an additive residual. The residual is computed by subtracting the content-only attention output from the style-content attention output and then adding the scaled difference to the base-content representation. This way allows us to preserve the original content trajectory while injecting only the style-related residual information.

\subsection{Timestep-wise Uncertainty Analysis}
\label{app:E2}
\begin{figure}[H]
\vspace{-15pt}
    \centering
    \includegraphics[width=\linewidth]{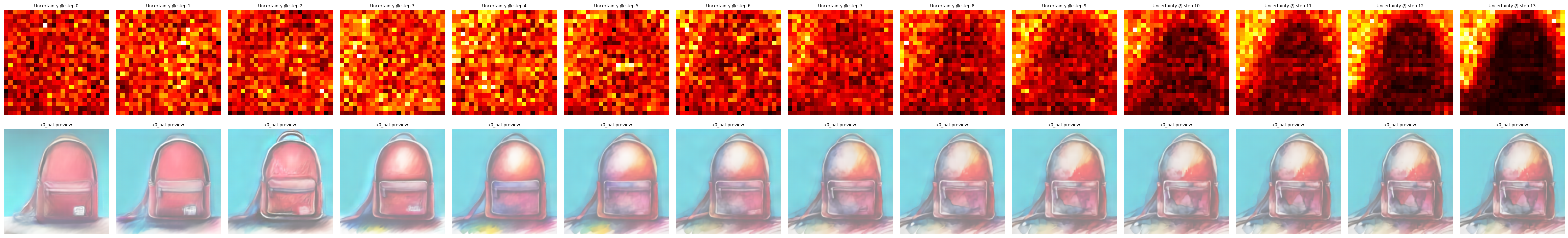}
    \includegraphics[width=\linewidth]{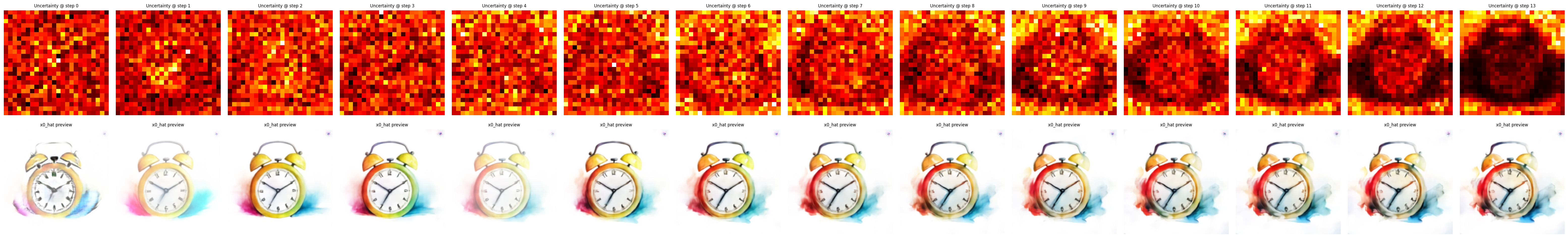}
    \includegraphics[width=\linewidth]{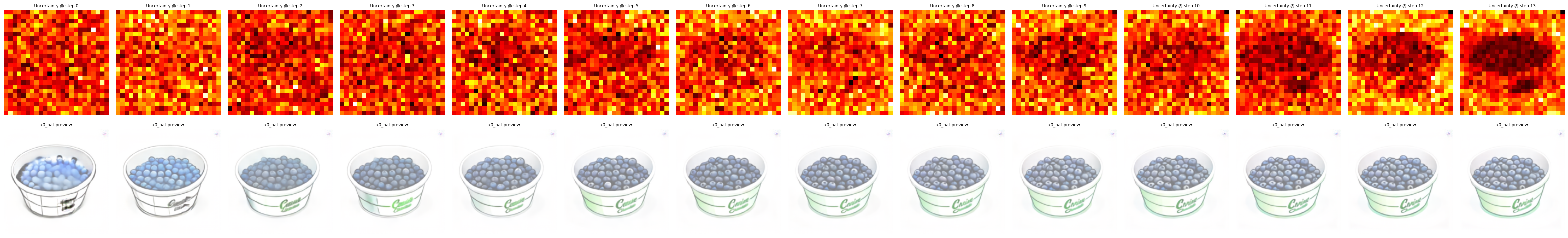}
    \caption{
    Timestep-wise uncertainty visualization during reverse diffusion. 
    For each example, the top row shows the estimated uncertainty map at different denoising steps, while the bottom row shows the corresponding predicted clean-latent preview $\widetilde{x}_{0|t}$.
    }
    \label{fig:timestep_uncertainty}
\vspace{-20pt}
\end{figure}

\Cref{fig:timestep_uncertainty} visualizes the estimated uncertainty maps across denoising steps. At early timesteps, the predicted clean preview $\widetilde{x}_{0|t}$ is still ambiguous, and the uncertainty is high and spatially diffuse. As denoising progresses, the object structure becomes clearer and the uncertainty gradually decreases. In particular, the regions corresponding to the main object, where the style has been injected, become more stable in the later timesteps. This can be observed from the uncertainty maps becoming less noisy and more concentrated, while the predicted clean previews show a more consistent object shape and appearance. These results indicate that uncertainty provides a useful reliability signal for adaptive guidance: stronger intervention is applied when the generation is unstable, while weaker intervention is used once the stylized object becomes stable.

\subsection{Analysis of the Starting Noise Step}
\label{app:E3}

\begin{figure}[H]
\vspace{-15pt}
    \centering
    \includegraphics[width=\linewidth]{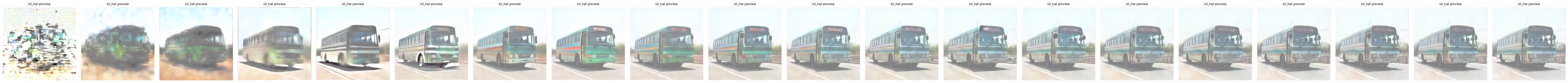}
    \includegraphics[width=\linewidth]{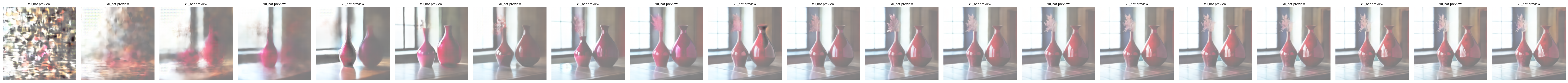}
    \includegraphics[width=\linewidth]{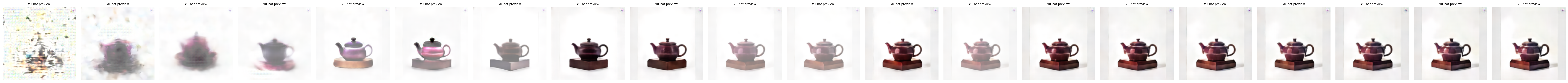}
    \includegraphics[width=\linewidth]{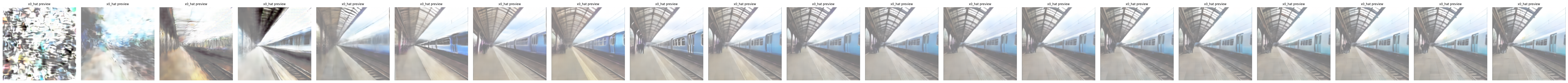}
    \caption{
Analysis of the starting noise step. Each row shows the predicted clean-latent previews $\widetilde{x}_{0|t}$ along the reverse diffusion trajectory for one example. Early high-noise steps produce ambiguous structures, while later steps become increasingly stable and content-specific.
}
\label{fig:starting_noise_step}
\vspace{-20pt}
\end{figure}

\Cref{fig:starting_noise_step} visualizes the predicted clean-latent previews $\widetilde{x}_{0|t}$ across denoising steps. At early high-noise steps, the previews are highly ambiguous and the object structure is not yet reliable. Starting from this region provides more freedom for style transfer, but it also increases the risk of changing the content layout or object identity. In contrast, very late steps already contain a stable content structure, which helps preserve the original image but leaves limited room for injecting the target style. Therefore, we choose an intermediate starting noise level $\rho=0.7$, which avoids the most unstable early region while still keeping enough flexibility for stylization.

\subsection{Uncertainty Sample Sensitivity and Computational Cost}
\label{app:E4}

We further analyze the effect of the number of perturbations $M$ on generation quality and computational cost. For each denoising step, the $M$ perturbed samples are processed jointly in a single batch. Therefore, uncertainty estimation introduces only one additional denoising-network forward pass per timestep, independent of $M$. Consequently, TRACE requires 28 NFE compared with 14 NFE for constant guidance, while increasing $M$ mainly affects runtime and memory through the larger batch size.

\begin{table}[H]
\vspace{-15pt}
\centering
\caption{Effect of uncertainty samples $M$ on generation quality and computational cost over 8 styles and 13 content images. Runtime is reported as mean $\pm$ standard deviation over 30 runs. The experiments are conducted on an A100 GPU 40GB.}
\label{tab:uncertainty_m_efficiency}

\setlength{\tabcolsep}{4pt}
\renewcommand{\arraystretch}{1.08}

\resizebox{\columnwidth}{!}{%
\begin{tabular}{l c ccc ccc}
\toprule
& &
\multicolumn{3}{c}{\textbf{Generation Quality}}
& \multicolumn{3}{c}{\textbf{Efficiency}} \\
\cmidrule(lr){3-5}
\cmidrule(lr){6-8}

\textbf{Method}
& \textbf{$M$}
& \textbf{CSD}$\uparrow$
& \textbf{DINO}$\uparrow$
& \textbf{DCL}$\downarrow$
& \textbf{Time (s)}$\downarrow$
& \textbf{VRAM (GB)}$\downarrow$
& \textbf{NFE}$\downarrow$ \\
\midrule

Const
& --
& 32.386
& 66.689
& 20.913
& \textbf{15.448 $\pm$ 0.180}
& 18.499
& \textbf{14}
\\

\midrule

TRACE
& 2
& 32.292
& \textbf{70.713}
& 18.838
& 31.019 $\pm$ 0.441
& \textbf{18.486}
& 28
\\

\ours TRACE
& \textbf{4}
& \textbf{37.394}
& 67.582
& \textbf{17.975}
& 33.923 $\pm$ 0.386
& 18.707
& 28
\\

TRACE
& 8
& 34.883
& 70.241
& 18.890
& 41.682 $\pm$ 0.145
& 20.061
& 28
\\

\bottomrule
\end{tabular}%
}
\vspace{-15pt}
\end{table}

As shown in \cref{tab:uncertainty_m_efficiency}, increasing $M$ does not increase NFE because all perturbations are evaluated in parallel. At $M=4$, TRACE requires $33.923$\,s and $18.707$\,GB while achieving the highest CSD and lowest DCL among the tested settings. We therefore use $M=4$ as the default setting.

\subsection{Adaptive Guidance vs. Handcrafted Schedules}
\label{app:E5}
\begin{figure}[H]
\vspace{-20pt}
    \centering
    \includegraphics[
        width=\columnwidth
    ]{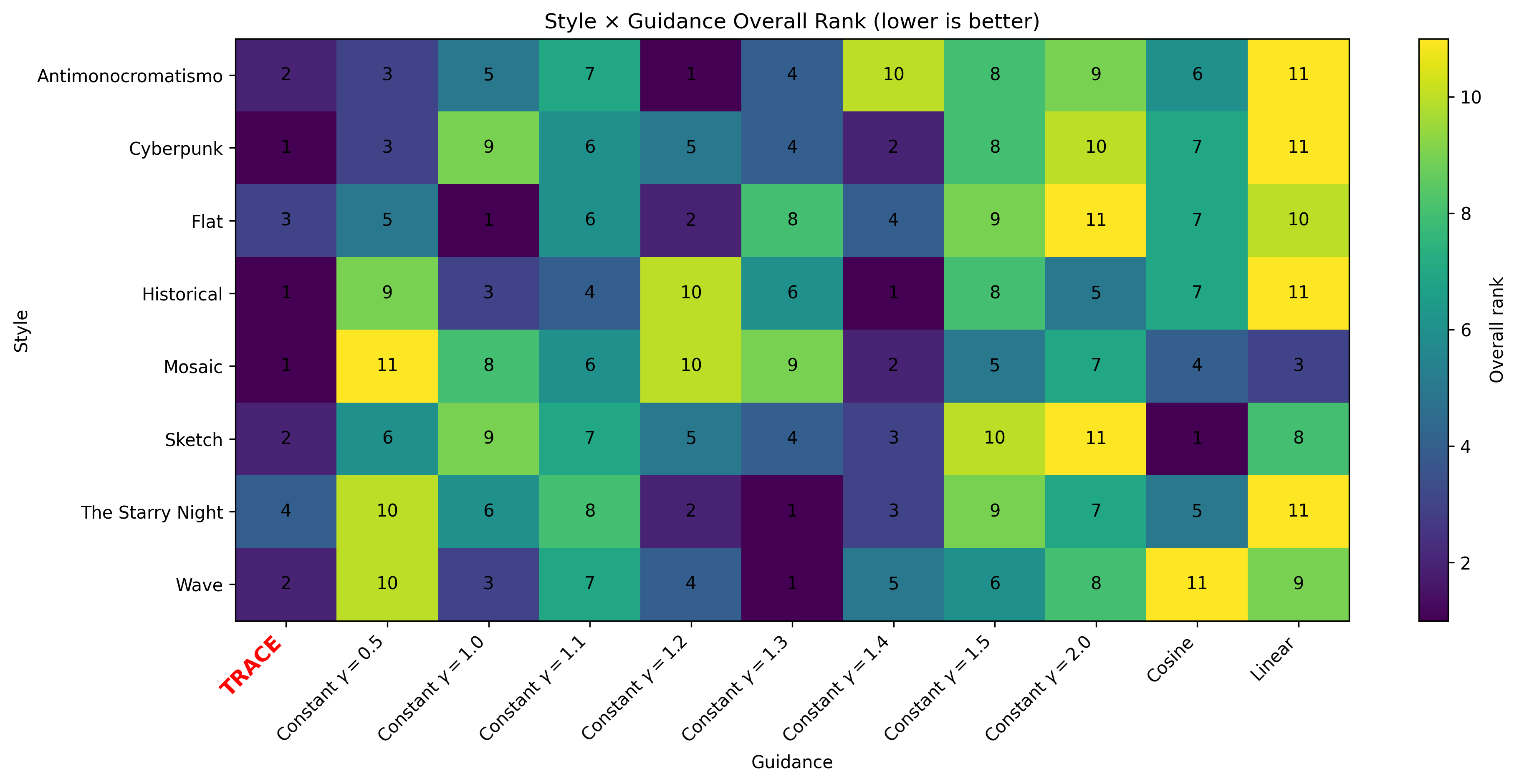}
    \caption{
    Cross-style ranking of fixed and handcrafted guidance configurations.
    }
    \label{fig:cross_style_ranking}
    \vspace{-20pt}
\end{figure}

\paragraph{\textbf{Cross-style ranking analysis.}}
To examine whether a single handcrafted guidance configuration generalizes across styles, we rank fixed, linear, and cosine schedules for each style according to their generation-quality trade-offs. As shown in \cref{fig:cross_style_ranking}, the ranking varies substantially across styles, indicating that a globally fixed schedule is difficult to tune consistently. In contrast, TRACE adapts its guidance from the uncertainty of the current denoising state without requiring style-specific schedule selection.

\begin{figure}[H]
\vspace{-20pt}
    \centering
    \includegraphics[
        width=\columnwidth
    ]{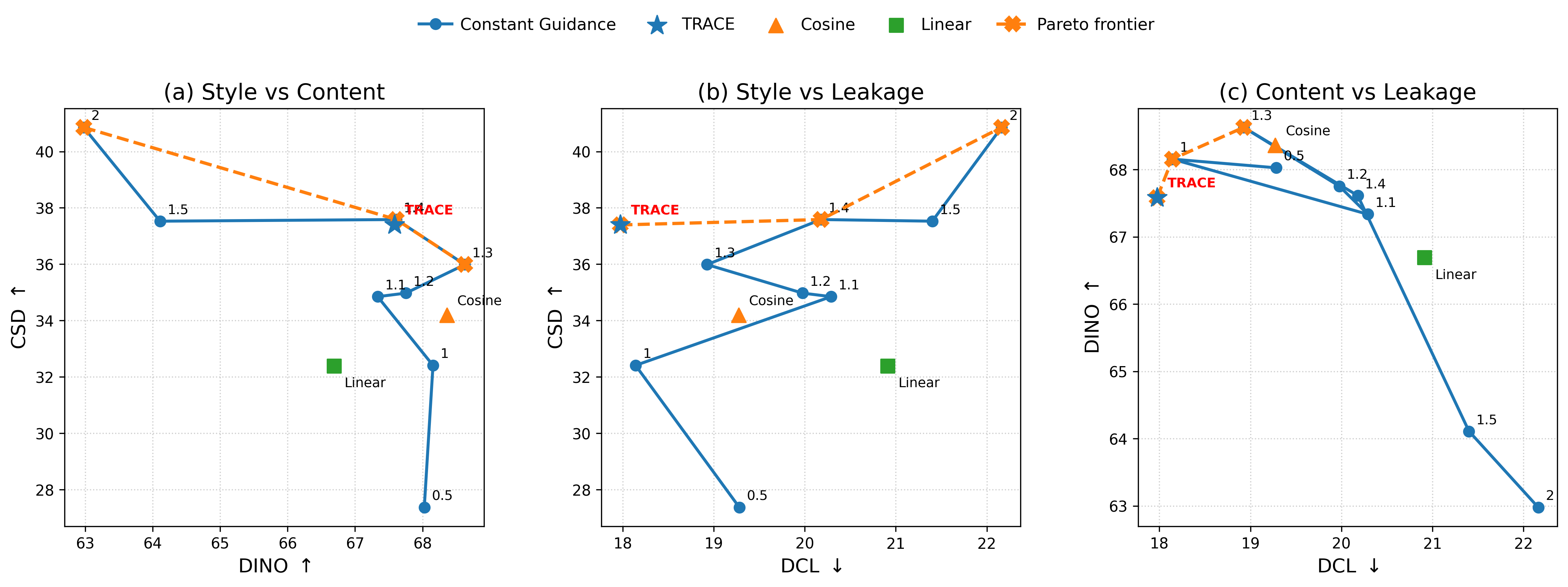}
    \caption{
    Trade-offs of fixed, linear, cosine, and uncertainty-adaptive guidance under different guidance strengths.
    }
    \label{fig:guidance_tradeoff}
    \vspace{-20pt}
\end{figure}

\paragraph{\textbf{Guidance trade-off analysis.}}
We further compare fixed, linear, and cosine schedules under different guidance strengths. As shown in \cref{fig:guidance_tradeoff}, their preferred operating points vary across schedules and styles, whereas TRACE adaptively modulates the guidance along the denoising trajectory.

\begin{figure}[H]
\vspace{-20pt}
    \centering
    \includegraphics[
        width=\columnwidth
    ]{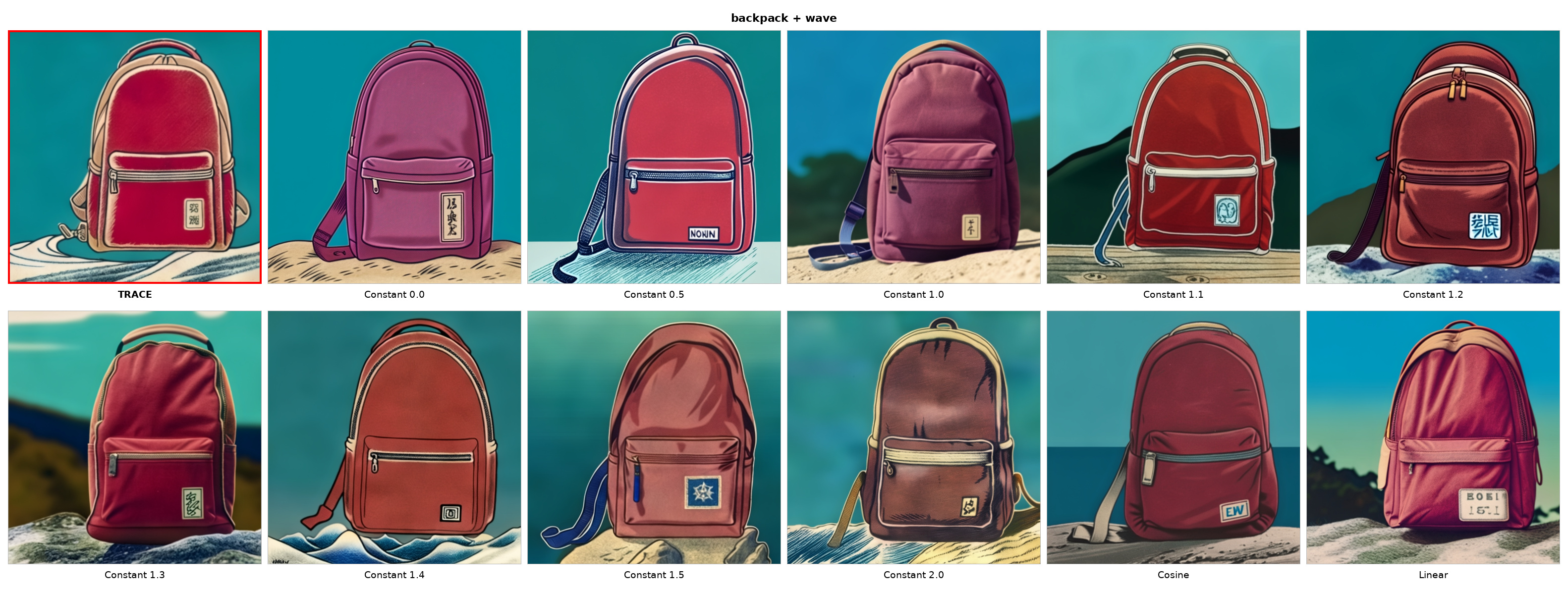}
    \caption{Qualitative comparison of uncertainty-adaptive, constant, and handcrafted time-dependent guidance.
    }
    \label{fig:adaptive_guidance_qualitative}
    \vspace{-20pt}
\end{figure}

\paragraph{\textbf{Qualitative comparison.}}
The qualitative results in \cref{fig:adaptive_guidance_qualitative}
show that uncertainty-adaptive guidance produces fewer visible artifacts than constant and handcrafted time-dependent schedules in the shown examples.

\subsection{Style Disentanglement and Held-out Style Generalization}
\label{app:E6}

We further evaluate whether the learned content-style subspaces capture reusable style directions rather than memorizing the style families used during offline decomposition. We additionally construct a style-disentanglement setting in which reference images share the same visual style but contain different semantic content.

\begin{table}[H]
\vspace{-15pt}
\centering
\caption{
Generalization and style-disentanglement analysis. Held-out evaluation uses 6 styles $\times$ 13 contents, while style disentanglement uses 4 styles with 3 references each $\times$ 13 contents.
}
\label{tab:generalization_disentanglement}

\setlength{\tabcolsep}{3.2pt}
\renewcommand{\arraystretch}{1.08}

\resizebox{\columnwidth}{!}{%
\begin{tabular}{l cc ccc c}

\toprule
&
\multicolumn{2}{c}{\textbf{Style}}
&
\multicolumn{3}{c}{\textbf{Content}}
&
\multicolumn{1}{c}{\textbf{Leakage}}
\\

\cmidrule(lr){2-3}
\cmidrule(lr){4-6}
\cmidrule(lr){7-7}

\textbf{Setting / Method}
& \textbf{CSD}$\uparrow$
& \textbf{SRA}$\uparrow$
& \textbf{DINO}$\uparrow$
& \textbf{CLIP-I}$\uparrow$
& \textbf{LPIPS}$\downarrow$
& \textbf{DCL}$\downarrow$
\\

\midrule
\multicolumn{7}{l}{
\textbf{Held-out Style-Family Generalization}
}
\\
\midrule

Full-data
& 43.213
& 96.875
& 68.909
& \textbf{81.269}
& \textbf{36.040}
& 19.333
\\

\ours Held-out Style Family
& \textbf{43.292}
& \textbf{96.875}
& \textbf{69.552}
& 80.684
& 36.128
& \textbf{19.162}
\\

\midrule
\multicolumn{7}{l}{
\textbf{Style Disentanglement}
}
\\
\midrule

No Decomposition
& \textbf{62.819}
& \textbf{66.333}
& 55.170
& 75.449
& 38.674
& 36.307
\\

\ours TRACE
& \underline{46.256}
& 63.333
& \underline{71.374}
& \underline{84.238}
& 32.643
& \underline{20.301}
\\

\addlinespace[2pt]

Rb-Modulation
& 34.785
& 43.333
& \textbf{78.369}
& \textbf{91.382}
& \textbf{27.427}
& \textbf{15.170}
\\

AttenST
& 38.906
& 58.333
& 65.869
& 81.762
& \underline{30.048}
& 23.923
\\

StyleID
& 37.798
& \underline{65.000}
& 67.564
& 84.073
& 32.220
& 22.218
\\

\bottomrule
\end{tabular}%
}
\vspace{-15pt}
\end{table}

\paragraph{\textbf{Style disentanglement.}}
We first evaluate whether the proposed content-style decomposition can suppress semantic information carried by the style reference. To this end, we construct a same-style, different-semantics setting, where multiple reference images share a similar visual style but depict different semantic content. As shown in \cref{tab:generalization_disentanglement}, removing the decomposition substantially degrades content preservation and increases semantic leakage. In particular, TRACE improves DINO from $55.170$ to $71.374$ and reduces DCL from $36.307$ to $20.301$ compared with the variant without decomposition, while retaining higher CSD than AttenST and StyleID. This indicates that the decomposition reduces reference-induced semantic leakage while preserving the desired visual style.

\begin{figure}[H]
\vspace{-20pt}
    \centering
    \includegraphics[
        width=\columnwidth
    ]{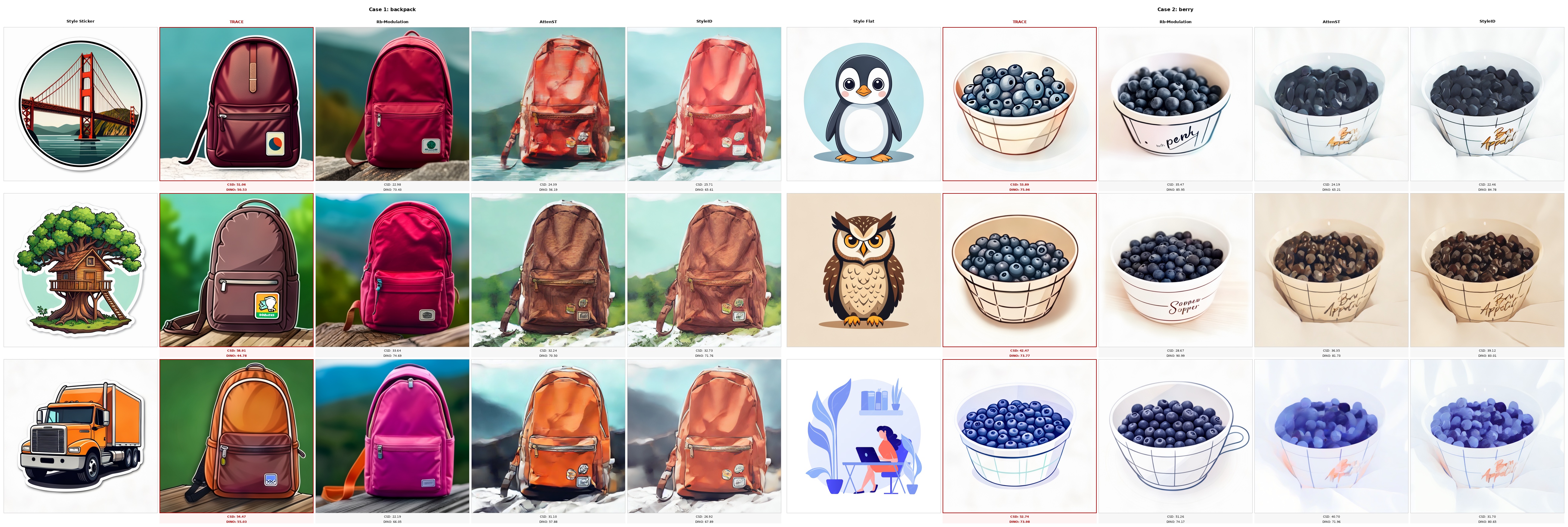}
    \caption{Style disentanglement under semantically different references sharing the same visual style.
    }
    \label{fig:style_disentanglement_qualitative}
    \vspace{-20pt}
\end{figure}

\paragraph{\textbf{Held-out style-family generalization.}}
We further examine whether the learned content-style subspaces generalize beyond the style families used during offline decomposition. For this evaluation, the target style family is excluded from the offline subspace construction and used only at inference time. As shown in \cref{tab:generalization_disentanglement}, performance remains nearly unchanged when the target style family is held out: CSD changes from $43.213$ to $43.292$, DINO from $68.909$ to $69.552$, and DCL from $19.333$ to $19.162$. These results suggest that the learned subspaces capture reusable content and style directions rather than relying on the presence of the target style family during offline estimation.

\begin{figure}[H]
\vspace{-20pt}
    \centering
    \includegraphics[
        width=0.85\columnwidth
    ]{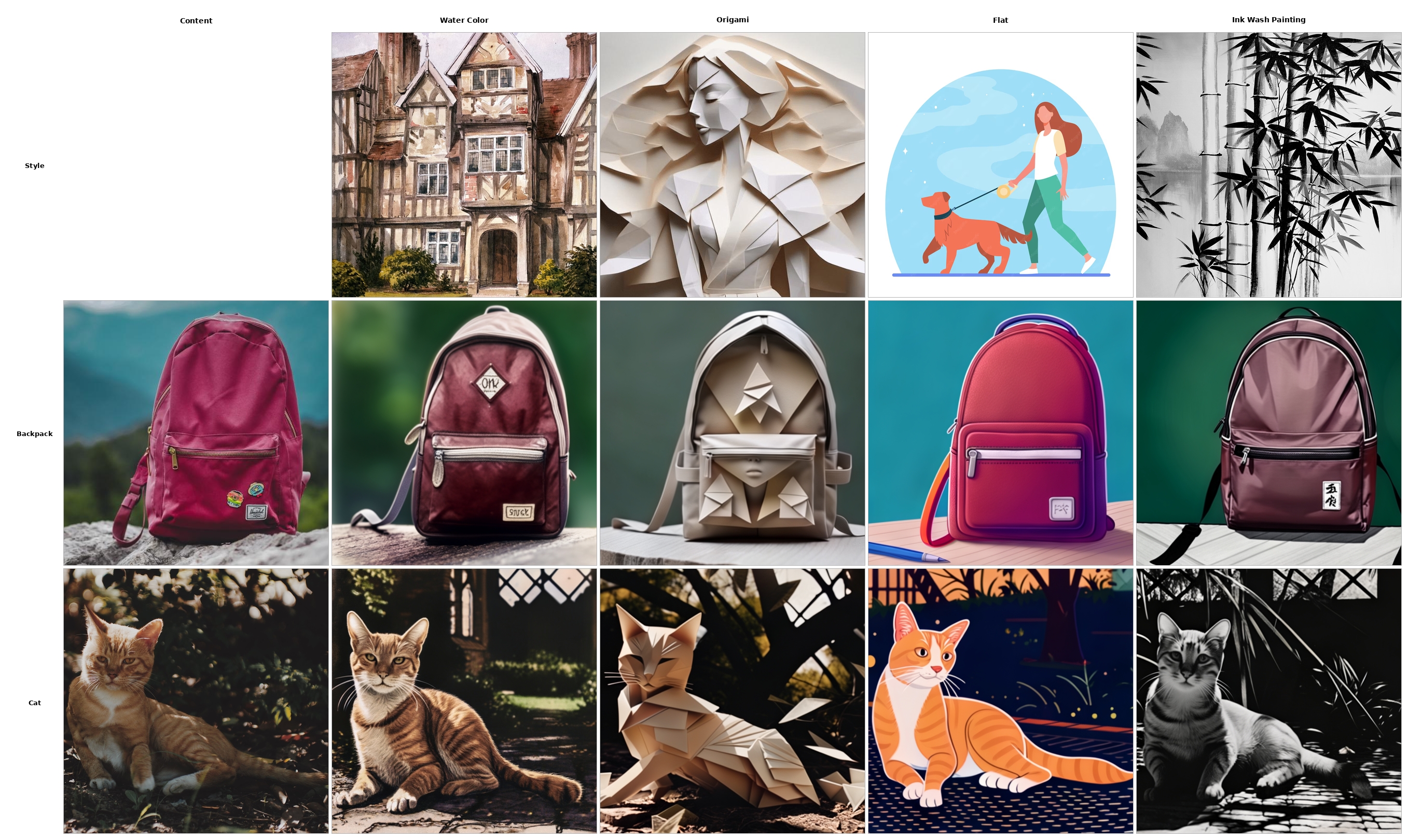}
    \caption{
    Qualitative result on an unseen origami-style reference.
    }
    \label{fig:unseen_style}
\vspace{-20pt}
\end{figure}

\subsection{Trade-off Analysis}
\label{app:E7}

\begin{figure}[H]
\vspace{-20pt}
    \centering
    \includegraphics[
        width=\columnwidth
    ]{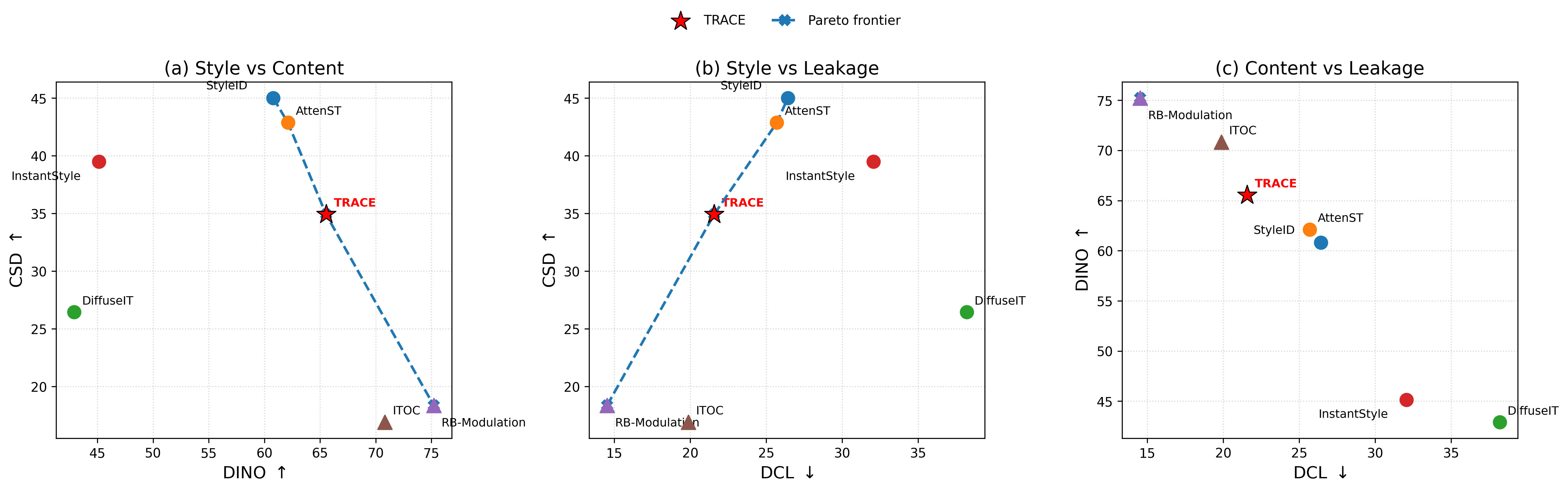}
    \caption{
    Pairwise Pareto trade-offs among representative style-transfer methods.
    }
    \label{fig:baseline_pareto}
\vspace{-15pt}
\end{figure}

Style transfer involves competing objectives among style fidelity, content preservation, and semantic leakage. We therefore examine pairwise Pareto trade-offs among representative methods rather than relying on a single aggregate criterion. As shown in \cref{fig:baseline_pareto}, different methods occupy different regions of the trade-off space, highlighting the balance achieved by TRACE across the competing objectives.

\subsection{Additional Qualitative Results}
\label{app:E8}

\begin{figure}[H]
\vspace{-20pt}
    \centering
    \includegraphics[width=\linewidth]{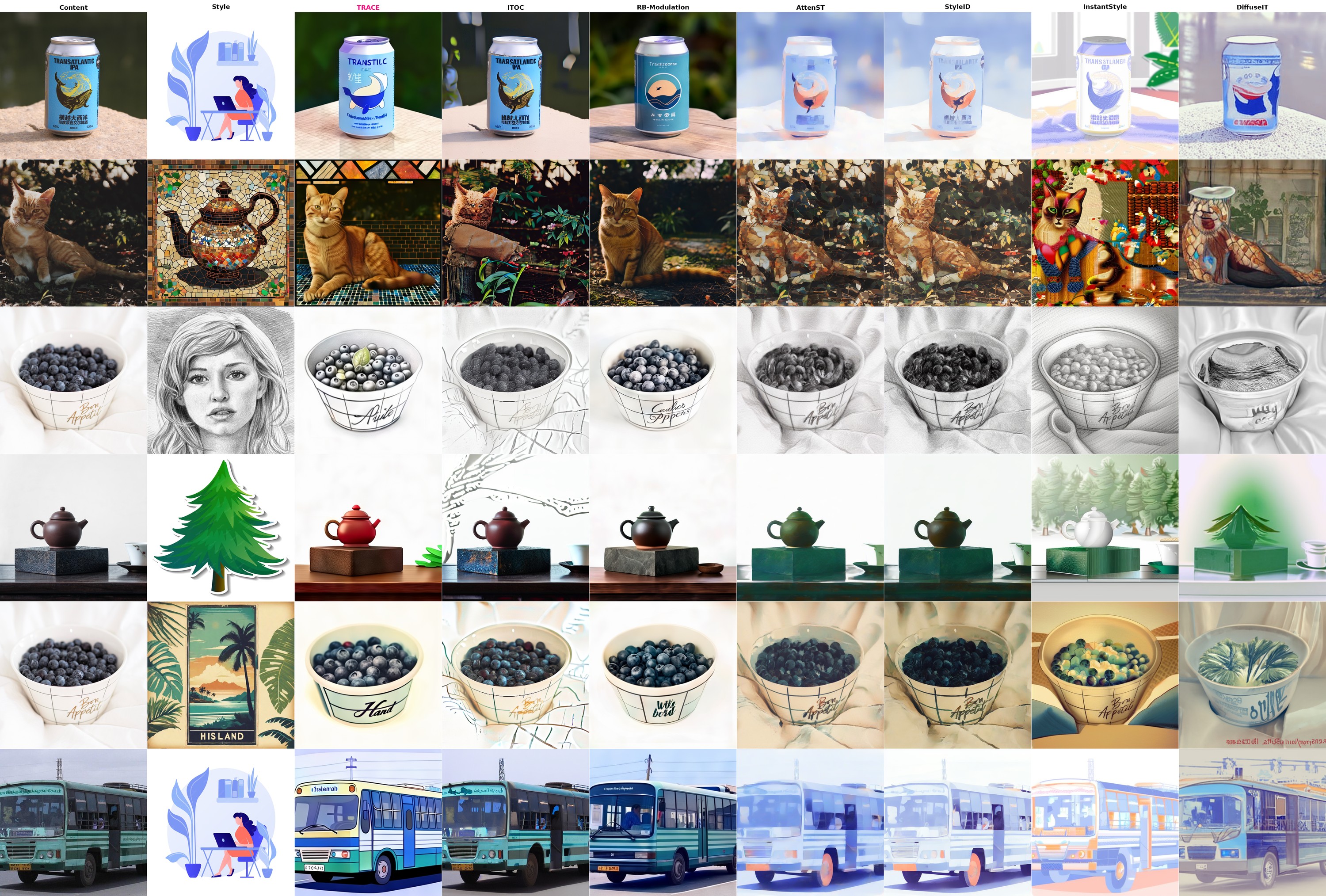}
    \caption{
    Additional qualitative comparison with baseline methods using a texture-dominant style reference. TRACE transfers local texture patterns while better preserving the main content structure.
    }
    \label{fig:comparison_texture}
    \vspace{-15pt}
\end{figure}

\begin{figure}[H]
    \vspace{-15pt}
    \centering
    \includegraphics[width=\linewidth]{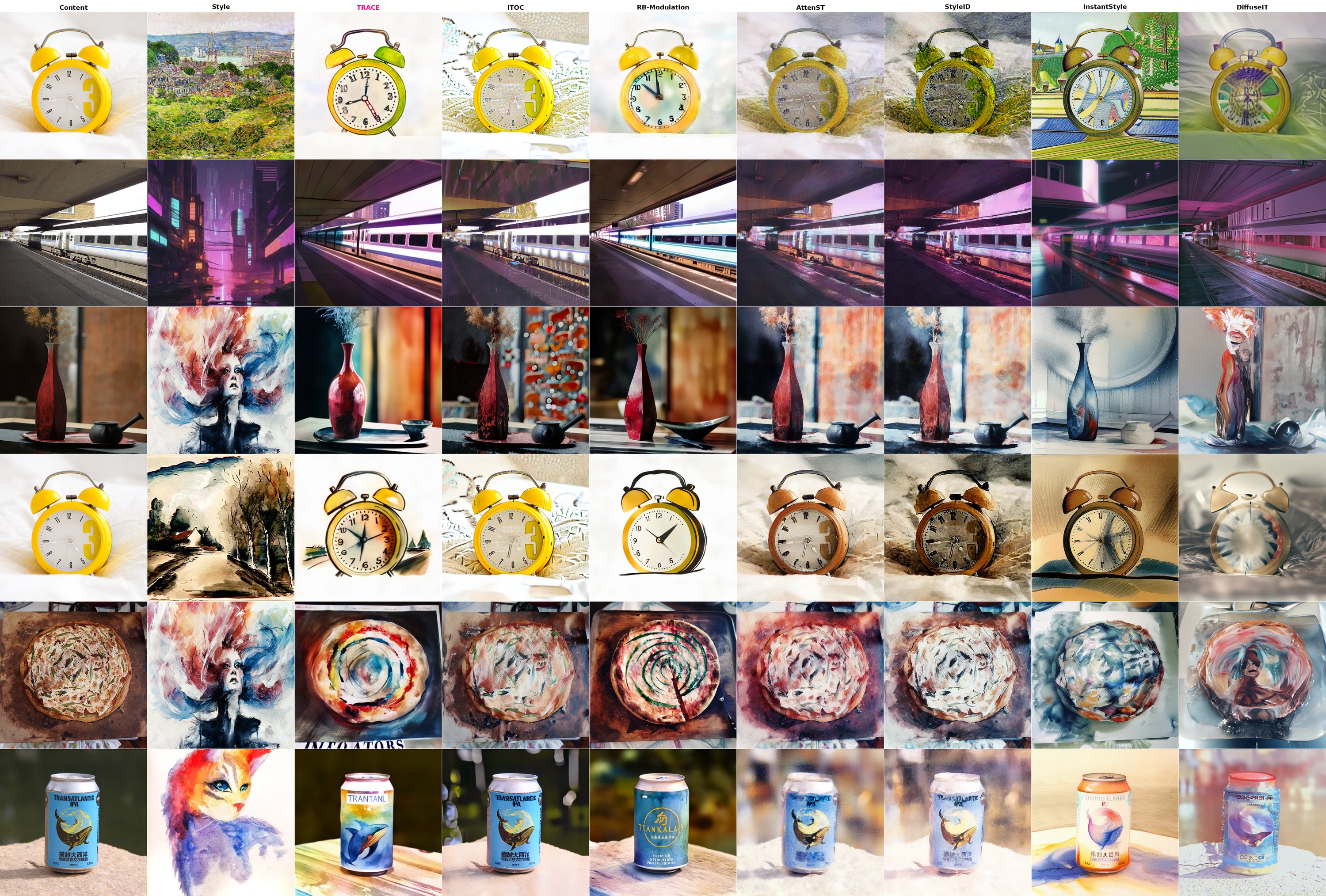}
    \caption{
    Additional qualitative comparison with baseline methods using a color-dominant style reference. TRACE captures the target color distribution while maintaining the semantic structure of the content image.
    }
    \label{fig:comparison_color}
        \vspace{-15pt}
\end{figure}

\paragraph{\textbf{Additional qualitative Comparison.}}
\Cref{fig:comparison_texture} and \cref{fig:comparison_color} provide additional qualitative comparisons under different style characteristics. For the texture-dominant style reference, baseline methods may transfer strong local patterns but can also introduce structural distortion or semantic leakage. For the color-dominant style reference, some baselines capture the global color distribution but may weaken content identity or object boundaries. In contrast, TRACE better preserves the main content structure while still transferring the target visual appearance, showing a more balanced style-content trade-off.

\begin{figure}[H]
    \centering
    \includegraphics[width=\linewidth]{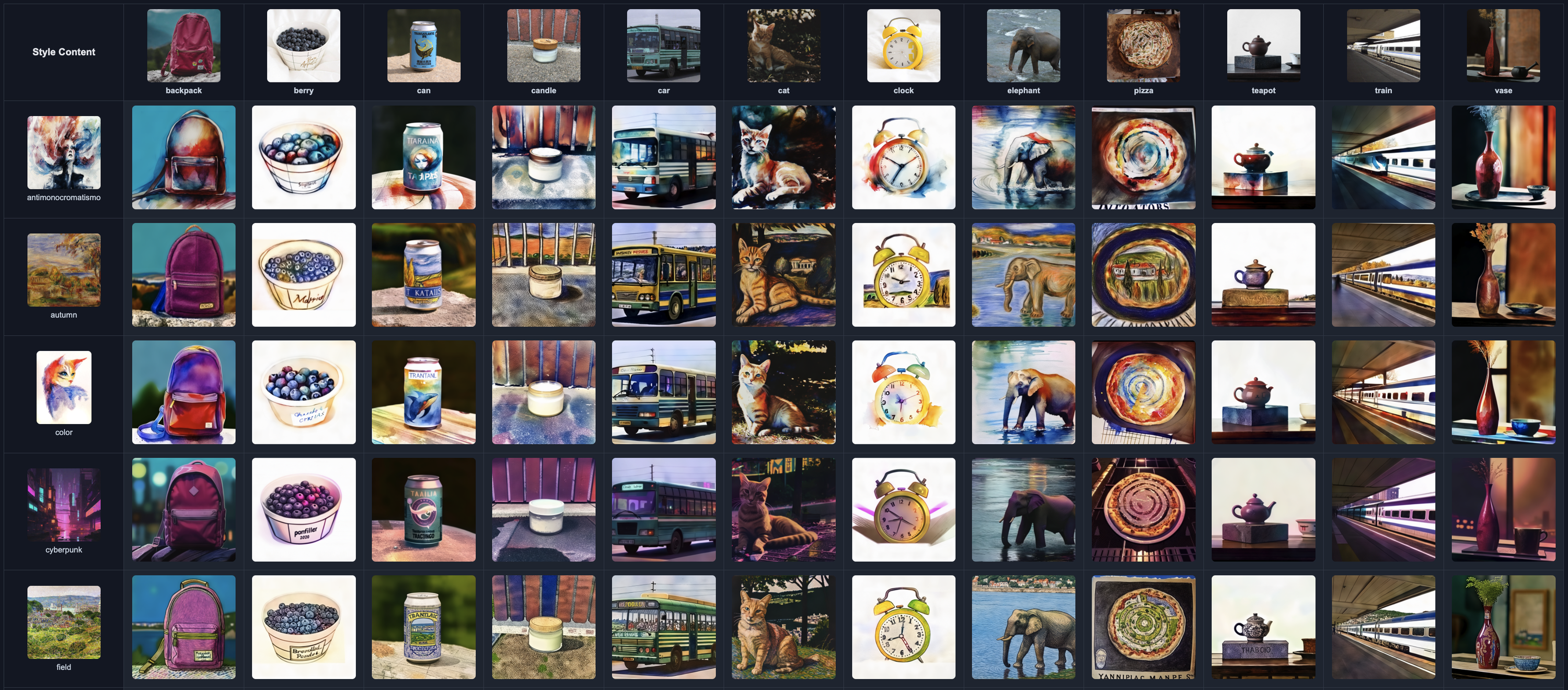}
    \caption{Representative stylization results of TRACE on different content and style pairs. Each example shows how the target style is transferred while preserving the main content structure.
    }
    \label{fig:trace_additional_samples}
\end{figure}

To further illustrate the qualitative performance of TRACE, we provide additional stylization results on diverse content-style pairs in \cref{fig:trace_additional_samples}.

\paragraph{\textbf{Background preservation.}}
Our evaluation primarily uses StyleAligned and DreamBooth images, which are predominantly object-centric. Therefore, our preservation claims mainly concern foreground structure and semantic content rather than strict background fidelity. In some examples, strong stylization may still alter or wash out background appearance.

\end{document}